\pdfoutput=1

\documentclass[11pt]{article}
\usepackage{authblk}
\usepackage[]{ACL2023}

\usepackage{times}
\usepackage{latexsym}
\usepackage{graphicx}
\usepackage{multirow}
\usepackage{multicol}
\usepackage{amsmath}
\usepackage{subfigure}
\usepackage{float}
\usepackage{soul}
\usepackage{hyperref}

\usepackage{subcaption} 
\usepackage[T1]{fontenc}

\usepackage[utf8]{inputenc}

\usepackage{microtype}
\usepackage{amsmath}

\usepackage{inconsolata}

\title{Pretraining Language Models Using Translationese}

\author[1]{\bf Meet Doshi}
\author[2]{\bf Raj Dabre}
\author[1]{\bf Pushpak Bhattacharyya}

\affil[1]{CFILT, Indian Institute of Technology Bombay, Mumbai, India}
\affil[2]{National Institute of Information and Communications Technology, Kyoto, Japan}
\affil[2]{IIT Madras, Chennai, India}
\affil[1]{\texttt{\{meetdoshi,pb\}@cse.iitb.ac.in}}
\affil[2]{{\texttt{raj.dabre@nict.go.jp}}}

\begin{document}
\maketitle
\begin{abstract}

In this paper, we explore the utility of \textit{Translationese} as synthetic data created using machine translation for pre-training language models (LMs) for low-resource languages (LRLs). Our simple methodology consists of translating large amounts of web-crawled monolingual documents (\textit{clean}) into the LRLs, followed by filtering the translated documents using tiny LMs trained on small but \textit{clean} LRL data. Taking the case of Indian languages, we pre-train LMs from scratch with 28M and 85M parameters, and then fine-tune them for 5 downstream natural language understanding (NLU) and 4 generative (NLG) tasks. We observe that pre-training on \textit{filtered synthetic} data leads to relative performance drops of only 0.87\% for NLU and 2.35\% for NLG, compared to pre-training on \textit{clean} data, and this gap further diminishes upon the inclusion of a small amount of \textit{clean} data. We also study the impact of \textit{synthetic} data filtering and the choice of source language for \textit{synthetic} data generation. Furthermore, evaluating continually pre-trained larger models like Gemma-2B and Llama-3-8B in few-shot settings, we observe that using \textit{synthetic} data is competitive with using \textit{clean} data.
Our findings suggest that \textit{synthetic} data shows promise for bridging the pre-training gap between English and LRLs.
\end{abstract}

\section{Introduction}
Large language models (LLMs) \cite{NEURIPS2020_1457c0d6,workshop2022bloom,almazrouei2023falcon,lin-etal-2022-shot} have been able to perform very well on downstream tasks like MMLU \cite{DBLP:conf/iclr/HendrycksBBZMSS21}, Big-Bench \cite{DBLP:journals/corr/abs-2206-04615}, etc, and have even started to reach human potential in many of these tasks. But this performance has very largely been credited to their scale and the vast amount of data that they have been fed. Most of these language models (LMs) perform well in languages like English where abundant data is available \cite{kudugunta2023madlad}, but a vast majority of languages don't have comparable data as compared to English. As a consequence, many LLMs, both monolingual and multilingual, involving these languages still show poor performance for various downstream tasks. For example, the largest open source multilingual model BLOOM \cite{workshop2022bloom} covers 46 natural languages spanning 9 language families, but the top 5 languages comprise 74.14\% of the data. Even for models like mT5 \citep{xue-etal-2021-mt5}, the top 10 of 107 languages account for more than 75.48\% of the training data. Despite the benefits of multilingualism \cite{10.1145/3406095}, this data skew still means that low-resource languages will underperform.


\begin{figure}[]
    \centering
    \includegraphics[width=70mm]{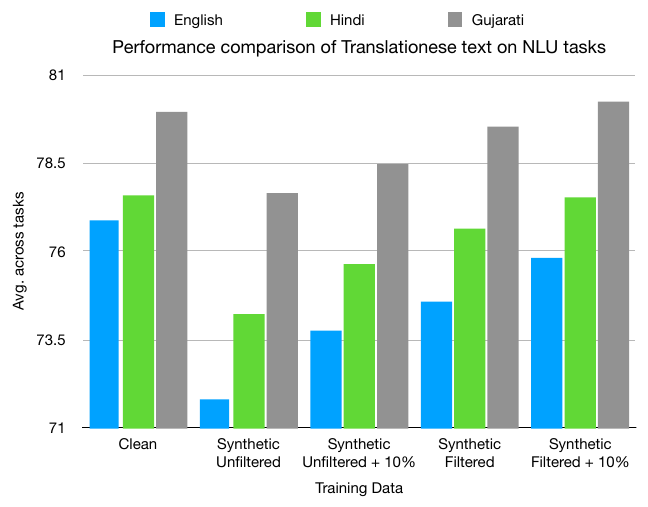}
    \caption{Comparing NLU performance in English, Hindi, and Gujarati shows that filtering synthetic data and adding 10\% clean data improves models, approaching the performance of those trained only on clean web data.}
    \label{fig:avg glue}
\end{figure}

Fortunately synthetic data is an option and previous works such as, but not limited to, back-translation \cite{sennrich-etal-2016-improving}, sequence distillation \cite{kim-rush-2016-sequence}, also known as forward translation, etc. have shown that synthetic data obtained using machine translation (MT) can supplement resource scarcity and can significantly enhance model performance \cite{popovic-etal-2020-neural,gala2023indictrans}. However, to the best of our knowledge, there has been no work on showing the effectiveness of synthetic data for pre-training LMs. Furthermore, the quality of synthetic data is also important, which many works take for granted. While round-trip-translation \cite{moon-etal-2020-revisiting} or referenceless neural quality estimation (QE) \cite{rei-etal-2021-references} are viable, they either involve twice the compute or a reasonably large model not available for most languages, and this might not be optimal to determine the quality of synthetic documents efficiently. We thus consider TinyLMs \cite{eldan2023tinystories} as an efficient alternative, which have been shown to model documents by their fluent paragraph generation capabilities. 

In this paper, we focus on Indic languages such as Hindi, Gujarati, and Marathi, and present a comprehensive study of the utility of \textit{synthetic} monolingual data, also called \textit{translationese} \cite{gellerstam1986translationese},  obtained using machine translation (MT) for pre-training LMs. We propose a simple framework that involves training tiny language models, henceforth TinyLMs, on original web-crawled data (clean) and then using them to filter synthetic data. We then compare LMs of different scales pre-trained on clean and synthetic data followed by fine-tuning on natural language understanding (NLP) and generation (NLG) downstream tasks, where we observe that, while unfiltered synthetic data based LMs are inferior compared to LMs trained on clean data, filtering leads to performance comparable to the latter. We further show that tuning these synthetic data LMs on small clean data leads to further improvements. We also show that these trends hold when continually pre-training LLMs such as Gemma-2B and Llama-3-8B.

\begin{figure*}[ht!]
    \centering
    \includegraphics[width=0.85\textwidth]{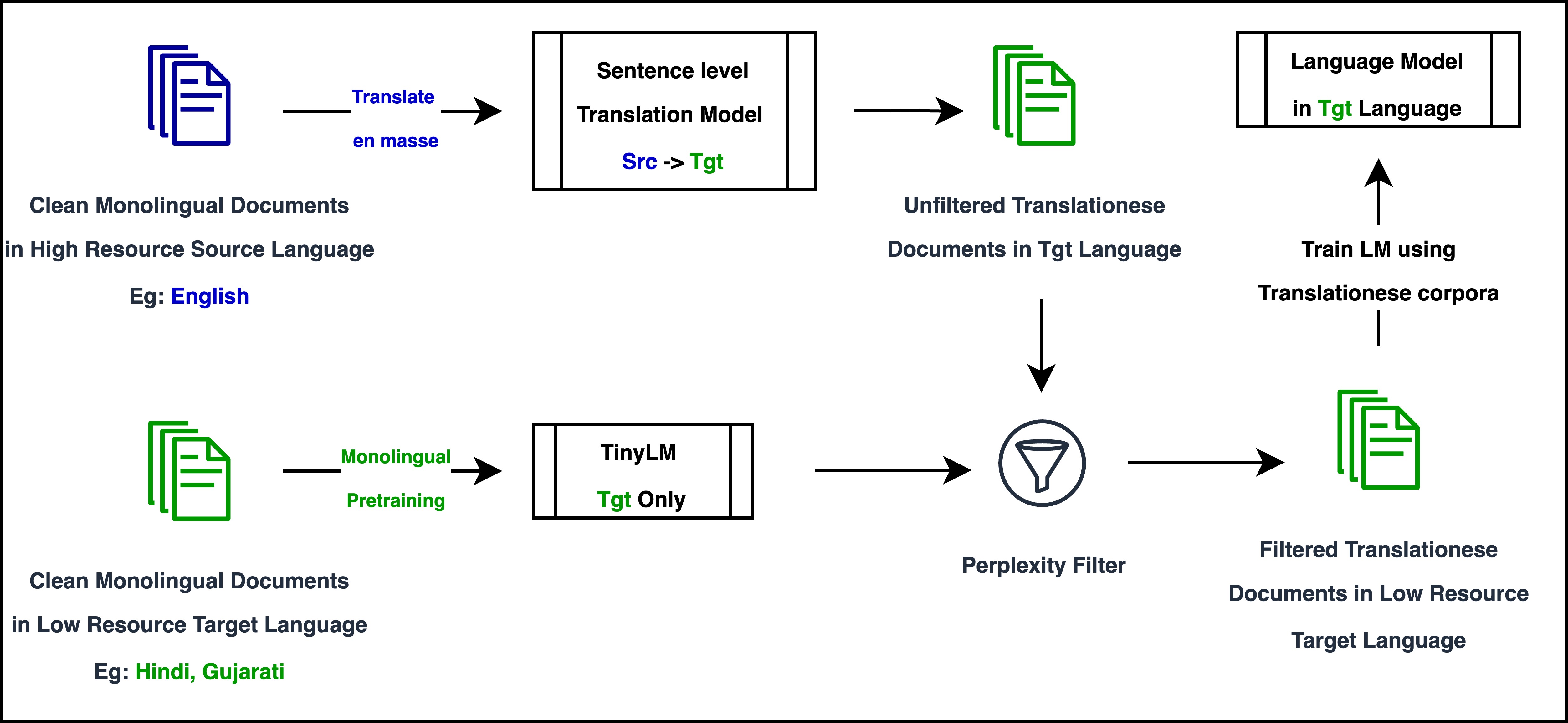}
    \centering
    \caption{Overview of our approach to pre-train language models using translationese data. We leverage rich monolingual corpora in the \textit{src} language and scarce corpora in the \textit{tgt} language. Our method involves employing a pre-trained machine translation model to translate \textit{src} to \textit{tgt}. We then filter, using perplexity, the resulting text using a TinyLM trained solely on clean \textit{tgt} monolingual data. The filtered synthetic data can be used to further pretrain larger language models.}
    \label{fig:filtering-translationese}
\end{figure*}

\noindent\textbf{Our contributions are:}

\noindent\textbf{a.} A simple framework involving high-quality MT models and TinyLMs trained on clean web-crawled data to mass-produce and filter synthetic data for LM training.

\noindent\textbf{b.} Demonstrating the efficacy of language models (up to Llama-3-8B) trained on filtered synthetic data across a range of NLU and NLG tasks for low resource Indic languages.

\noindent\textbf{c.} A new document-level monolingual corpora (\textit{IndicMonoDoc}) consisting of 39.5B tokens worth of monolingual clean document-level data spanning 22 scheduled languages and English\footnote{Our code and datasets are released at \href{https://github.com/meetdoshi90/TranslationesePretraining/}{https://github.com/meetdoshi90/TranslationesePretraining}}. 


\section{Related Work}
This paper focuses on creating, filtering, and utilizing synthetic data to train TinyLMs.

\noindent\textbf{Monolingual Data:} Previous efforts to collect monolingual corpora for Indic languages include the EMILLE/CIIL corpus \cite{mcenery2000emille}, HindMonoCorp \cite{bojar2014hindencorp}, Leipzig corpus \cite{goldhahn-etal-2012-building}, IndicCorpv1 \cite{kakwani2020indicnlpsuite}, and IndicCorpv2 \cite{doddapaneni2023towards}. While IndicCorpv2 is large, it is sentence-level and suitable for NLU models but not for longer contexts needed by language models. We extend these corpora and demonstrate the effectiveness of using synthetic data.

\noindent\textbf{Synthetic Data Generation and Quality Estimation:} Synthetic data aids NLP tasks like back translation for machine translation \cite{sennrich-etal-2016-improving, edunov-etal-2018-understanding, marie-etal-2020-tagged, DBLP:journals/corr/abs-1911-03362, ni-etal-2022-original} and native language identification \cite{goldin-etal-2018-native}. However, using synthetic data for pretraining LMs is less explored due to hallucination \cite{maynez-etal-2020-faithfulness} and ungrounded text \cite{thorne-etal-2018-fact}. Evaluation methods like RTT BLEU scores are computationally intensive, while others like BARTScore \cite{DBLP:conf/nips/YuanNL21}, T5Score \cite{qin-etal-2023-t5score}, MQM, and COMET \cite{rei-etal-2020-comet} require large-scale models or human annotations, limiting scalability. Approaches like KenLM \cite{heafield-2011-kenlm} have been used to filter monolingual corpora based on perplexity.

\noindent\textbf{Transfer Learning and Cross-Lingual Fine-Tuning:} Approaches like translate-train, as described by \citet{conneau-etal-2018-xnli}, involve fine-tuning a multilingual PLM using machine-translated training data and evaluating in the target language. \citet{oh-etal-2022-synergy} combined translate-train and translate-test for improved cross-lingual fine-tuning. In contrast, our work focuses on pretraining language models and exploring how synthetic text impacts pretraining and various downstream NLU and NLG tasks.

\noindent\textbf{TinyLMs:} Small LMs, even with 10M parameters, produce fluent and consistent text \cite{eldan2023tinystories}. Challenges like BabyLM \cite{warstadt-etal-2023-findings} focus on improving LMs within a fixed data budget. We take motivation from this and leverage TinyLMs for efficient filtering of synthetic documents.

\section{Methodology}
\label{sec: methodology}

In this section, we describe our framework for leveraging synthetic data for LM training. This process consists of collecting monolingual (\textit{clean}) data from the web for low-resource languages, training TinyLMs with it, translating \textit{clean} data from a high resource language such as English into low-resource languages, using the aforementioned TinyLMs to filter \textit{synthetic} data, and then using this filtered data to train LMs for downstream tasks. Our framework is illustrated in Figure~\ref{fig:filtering-translationese}.

\begin{figure}[h!]
    \centering
    \includegraphics[width=0.45\textwidth]{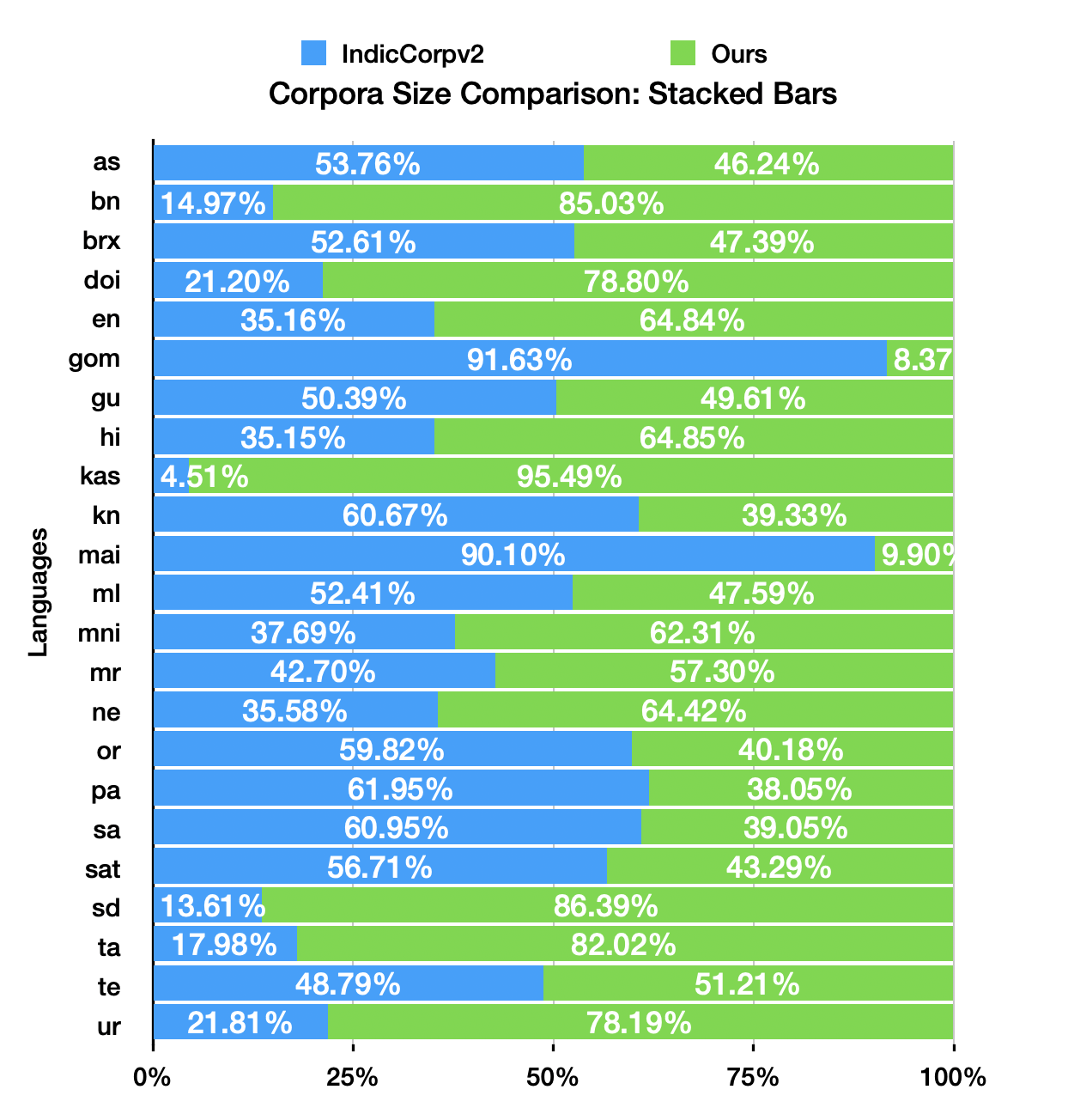}
    \caption{Language-wise corpora size comparison with IndicCorpv2 \cite{doddapaneni2023towards}: Stacked Bars}
    \label{fig:IndicMonoDoc}
\end{figure}

\subsection{Collecting Clean Monolingual Corpora}
\label{subsec: Indicmonodoc}

Following \citet{doddapaneni2023towards,rae2022scaling,nllbteam2022language}, for all languages of interest, we \textbf{a.} obtain a list of URLs to be crawled via word-level n-grams passed to a search engine, \textbf{b.} after URL deduplication, we crawl all applicable webpages, \textbf{c.} automatically and manually \cite{OrtizSuarezSagotRomary2019,2022arXiv220106642A} filter out unwanted text like HTML tags and emoticons, \textbf{d.} use language detection-based (LID) filtering using cld3\footnote{\url{https://github.com/google/cld3}} and IndicLID-FTN model \cite{madhani2023bhashaabhijnaanam} to discard languages not of interest, \textbf{e.} perform document filtering to drop documents containing offensive text using toxic words list provided by \citet{nllbteam2022language}, \textbf{f.} merge all the filtered corpus with Wikipedia, OSCAR \cite{OrtizSuarezSagotRomary2019} and some dumps of mC4 \cite{xue-etal-2021-mt5} and finally, \textbf{g.} perform deduplication at paragraph level using Murmurhash algorithm\footnote{\url{https://pypi.org/project/mmh3/}} with a 128-bit unsigned hash for each monolingual split of the corpora. 

We crawl data for English, with Indic context, and 22 Indic languages. As a result, we end up with IndicMonoDoc, with 27.5 billion tokens worth of Indic language documents and 12 billion tokens of English documents for a total of 39.5 billion tokens of \textit{clean} monolingual data. This is larger than the corpora released by \citet{doddapaneni2023towards}, surpassing it by almost 2 times. We use IndicMonoDoc for all experiments with \textit{clean} data. Figure~\ref{fig:IndicMonoDoc} gives an overview of the comparison of IndicMonoDoc. Note that, creation of IndicMonoDoc is important since IndicCorpV2 is a sentence-level corpus, and training LMs need a document-level corpus. It is important to note that we paid special attention to the low-resource languages. In this paper we only use data corresponding to English, Hindi, Gujarati and Marathi.
We report additional details of IndicMonoDoc in Appendix \ref{sec:indicmonodoc appendix}.



%

%


\subsection{Generating Translationese (Synthetic)}
We utilize state-of-the-art MT models like IndicTrans2 \cite{gala2023indictrans} to generate translationese data. Using beam search with a beam value of 5, we translate English tokens from the \textit{clean} corpus to the target languages. Due to token limits in MT models, we split documents using the Moses Sentence Splitter\footnote{\url{https://pypi.org/project/mosestokenizer/}} for sentence-level translations, then merge them back into documents. We use the 1B En-Indic version\footnote{\url{https://huggingface.co/ai4bharat/indictrans2-en-indic-1B}} of IndicTrans2 to translate 5B English tokens worth of documents from IndicMonoDoc into \textit{translationese} data for Hindi, Marathi and Gujarati.

\subsection{Tiny Language Models (TinyLMs)}
\label{sec:tinylms}

TinyLMs are small language models inspired by \citet{eldan2023tinystories}. We use the Transformer architecture \cite{vaswani2017attention} and train them with \textit{clean} monolingual documents. RoPE embeddings \cite{su2023roformer} are used instead of learned positional encodings for handling long documents. Following Chinchilla scaling laws \cite{hoffmann2022training}, we use compute-optimal word tokens. Although it is plausible to train a TinyLM on \textit{unfiltered translationese} data to filter itself, our preliminary experiments revealed that they favor poor-quality data and hence we avoid this route.


\subsection{Synthetic Data Filtering}
\label{sec:datafiltering}
We use these TinyLMs to filter the generated \textit{translationese} data. We do this by using perplexity as a measure of document quality score. For language models, perplexity quantifies how well a model predicts a sequence of tokens. A lower perplexity indicates a natural document. It is calculated by: 
$$\operatorname{PPL}(W)=\exp \left\{-\frac{1}{N} \sum_i^N \log p_\theta\left(w_i \mid w_{<i}\right)\right\}$$
where the negative log-likelihood measures the error of the model's predictions. While calculating perplexity over a sequence of tokens, $W \in$ \( w_1, w_2, \ldots, w_N \) we skip the first $s$ tokens where $s=10$, $e=1024$ and calculate loss until only the first $e$ tokens of the document. We find setting $e$ to larger values can lead to higher variance in the document scores due to the size of the TinyLM. Following initial analyses, we choose $s$ and $e$ such that we remove the high uncertainty of the language at the start of an unseen document and avoid penalizing longer documents due to the fragility of the extrapolation ability of TinyLM\footnote{During experiments we saw that these TinyLMs can only go up to a certain context length before deteriorating in quality.}. Note that it is important to choose $e$ such that the language model gives a uniform estimate of perplexity over an already seen sequence of tokens $\in$ {\( w_s, w_{s+1}, \ldots, w_e \)}. For our experiments, we use the TinyLMs to score all synthetically generated translationese data and calculate a document score using the above method. Following, \citet{laurenccon2022bigscience} we do subsampling by thresholding document perplexity scores except \citet{laurenccon2022bigscience} did it using Ken-LM \cite{heafield-2011-kenlm} and we do it using our TinyLM. We keep the threshold value such that we include enough documents to reach the computed optimal token count for pretraining experiments.

\section{Experiments}
\label{sec: experiments}
This section outlines the training procedures and datasets for the models described in Section \ref{sec: methodology}. We pre-train decoder only LMs and fine-tune all models from scratch in monolingual and bilingual settings using the causal language modeling (CLM) objective for NLG tasks and a linear classification head for classification tasks. We specify the dataset samples used for pretraining and fine-tuning, and analyze the effects of synthetic corpora on pretraining.
\subsection{Pretraining Data Settings}
We refer to translated text or translationese as \textbf{synthetic} or \textbf{syn} and original or web-crawled data as \textbf{clean} throughout our experiments. For the pretraining of all base models, we use the following naming convention to denote our training splits for each model:

\noindent \textbf{XX-clean:} This is a clean subset sampled randomly from IndicMonoDoc where XX represents the language English (EN), Hindi (HI) or Gujarati (GU).

\noindent \textbf{syn-XX\_yy-unfiltered:} Denotes synthetic monolingual documents in XX language generated by using yy as a source during translation.

\noindent \textbf{syn-XX\_yy-filtered:} Filtered synthetic data.

\noindent \textbf{+10\%:} Refers to extended pretraining on a cleaned subset of IndicMonoDoc with an additional 10\% tokens compared to regular training.

\noindent \textbf{BI-XX-YY Prefix:} Denotes bilingual models trained using an equal mixture of monolingual corpora in XX and YY languages. We append an \textit{\_syn} prefix to either XX or YY if a synthetic version of that language is employed in training, and a \textit{-parallel/nonparallel} tag to denote whether a parallel version of XX and YY are used or not.








Note, for each split we only use the number of tokens that are required to reach the point of optimality \cite{hoffmann2022training} by the language model. 

\subsection{Implementation and Training}

\noindent \textbf{Tokenizer}: We use a common byte-pair-encoding (BPE) \cite{DBLP:conf/acl/SennrichHB16a} tokenizer using Sentencepiece\footnote{\url{https://github.com/google/sentencepiece}} for all experiments. We train a shared vocabulary of 56k subwords between three languages, English, Hindi, and Gujarati by using 5 Million randomly sampled sentences per language and upsampling for Gujarati.

\noindent \textbf{TinyLMs:} We use Pytorch Lightning\footnote{\url{https://lightning.ai/docs/pytorch/stable/}} for our implementations and train TinyLMs as described in Section~\ref{sec:tinylms} for filtering. We use hidden sizes of 768 and have two variants, one with 4 layers (\textit{mini}) and one with 12 layers (\textit{base}; same as GPT2-base) with 28M and 85M non-embedding parameters respectively. The \textit{mini} models are trained on clean data with sequence lengths of 4096\footnote{We keep long sequence lengths to be able to handle long documents for filtering.} (\textit{mini-4k}) for filtering synthetic documents as described in Section~\ref{sec:datafiltering}. On the other hand, for our main pre-training and downstream fine-tuning experiments, we train \textit{mini} and \textit{base} models with sequence lengths of 1024 (\textit{mini-1k} and \textit{base-1k}). Following \citet{hoffmann2022training} we use 2.4 billion word tokens per language to compute optimal training of \textit{base} models. Since Gujarati has only 900M tokens in our dataset, whenever Gujarati is involved as the target, we train only the \textit{mini-1k} model. For models involving English and Hindi, we train both \textit{mini} and \textit{base} models. Additional details are in Appendix \ref{sec:training hyperparameters appendix}.

\subsection{Downstream Tasks and Evaluation}

We finetune the \textit{mini-1k} and \textit{base-1k} models for classification, regression, and generation tasks. Hyperparameter tuning is performed using the validation set for models trained only with clean data, and this process is repeated for different data splits. More details on hyperparameters and evaluation can be found in Appendix \ref{sec:training hyperparameters appendix}. Primary scores are reported on IndicGLUE \cite{kakwani2020indicnlpsuite} and IndicXNLI (iXNLI) \cite{aggarwal-etal-2022-indicxnli} for Hindi and Gujarati, and the GLUE benchmark validation set \cite{wang2018glue} for English. We also experiment with other generation tasks like CNN-Dailymail \cite{nallapati2016abstractive}, DailogSum \cite{chen-etal-2021-dialogsum}, XL-Sum \cite{hasan-etal-2021-xl}, IndicNLG \cite{kumar-etal-2022-indicnlg}, FLoRes-200 \cite{nllbteam2022language}, IN22-Conv \& IN22-Gen \cite{gala2023indictrans} and use standard evaluation metrics suitable for each task like accuracy, f1-score, Rouge-L \cite{lin-2004-rouge} and chrF++ \cite{popovic-2017-chrf}. Further details about each of the evaluation datasets can be found in Appendix \ref{subsec: evaluation datasets appendix}.

\begin{table*}[]
\centering
\subtable[Results on Hindi]{
\resizebox{\textwidth}{!}{%
\begin{tabular}{c|cccccc|ccccc}
\hline
{\color[HTML]{333333} } &
  \multicolumn{6}{c|}{{\color[HTML]{333333} \textbf{NLU}}} &
  \multicolumn{5}{c}{\textbf{NLG}} \\ \cline{2-12} 
\multirow{-2}{*}{{\color[HTML]{333333} \textbf{Model}}} &
  {\color[HTML]{333333} \textbf{iXNLI}} &
  \textbf{bbc-a} &
  \textbf{iitp-mr} &
  \textbf{iitp-pr} &
  \textbf{midas} &
  \multicolumn{1}{l|}{\textbf{Avg.}} &
  \textbf{\begin{tabular}[c]{@{}c@{}}Headline\\ Gen.\end{tabular}} &
  \textbf{\begin{tabular}[c]{@{}c@{}}Sentence\\ Summ.\end{tabular}} &
  \textbf{\begin{tabular}[c]{@{}c@{}}Question\\ Gen.\end{tabular}} &
  \textbf{Wikibio} &
  \multicolumn{1}{l}{\textbf{Avg.}} \\ \hline
{\color[HTML]{333333} HI-clean} &
  {\color[HTML]{333333} 73.61} &
  81.75 &
  72.58 &
  79.73 &
  80.34 &
  77.60 &
  27.54 &
  23.64 &
  24.84 &
  52.16 &
  32.04 \\ \hline
{\color[HTML]{333333} syn-HI\_en-unfiltered} &
  {\color[HTML]{333333} 72.87} &
  77.92 &
  64.36 &
  76.22 &
  79.91 &
  74.26 &
  27.29 &
  22.93 &
  24.22 &
  50.14 &
  31.14 \\
{\color[HTML]{333333} syn-HI\_en-unfiltered+10\%} &
  {\color[HTML]{333333} 74.63} &
  78.36 &
  67.75 &
  77.46 &
  80.17 &
  75.67 &
  26.98 &
  23.20 &
  24.76 &
  \textbf{51.34} &
  31.57 \\
{\color[HTML]{333333} syn-HI\_en-filtered} &
  {\color[HTML]{333333} \textbf{74.75}} &
  \textbf{81.06} &
  69.03 &
  78.58 &
  79.73 &
  76.63 &
  27.15 &
  23.10 &
  24.41 &
  49.88 &
  31.13 \\
{\color[HTML]{333333} syn-HI\_en-filtered+10\%} &
  {\color[HTML]{333333} 74.49} &
  80.94 &
  \textbf{71.61} &
  \textbf{79.92} &
  \textbf{80.64} &
  \textbf{77.52} &
  \textbf{27.87} &
  \textbf{24.23} &
  \textbf{24.87} &
  51.18 &
  \textbf{32.04} \\ \hline
\end{tabular}
}
}
\subtable[Results on Gujarati]{
\resizebox{0.75\textwidth}{!}{%
\begin{tabular}{c|ccc|cccc}
\hline
{\color[HTML]{333333} } &
  \multicolumn{3}{c|}{{\color[HTML]{333333} \textbf{NLU}}} &
  \multicolumn{4}{c}{\textbf{NLG}} \\ \cline{2-8} 
\multirow{-2}{*}{{\color[HTML]{333333} \textbf{Model}}} &
  {\color[HTML]{333333} \textbf{iXNLI}} &
  \textbf{iNLTK} &
  \multicolumn{1}{l|}{\textbf{Avg.}} &
  \textbf{\begin{tabular}[c]{@{}c@{}}Headline\\ Gen.\end{tabular}} &
  \textbf{\begin{tabular}[c]{@{}c@{}}Sentence\\ Summ.\end{tabular}} &
  \textbf{\begin{tabular}[c]{@{}c@{}}Question\\ Gen.\end{tabular}} &
  \multicolumn{1}{l}{\textbf{Avg.}} \\ \hline
{\color[HTML]{333333} GU-clean} &
  {\color[HTML]{333333} 67.8} &
  92.1 &
  79.95 &
  17.62 &
  13.82 &
  15.18 &
  15.54 \\ \hline
{\color[HTML]{333333} syn-GU\_en-unfiltered} &
  {\color[HTML]{333333} 65.51} &
  89.78 &
  77.65 &
  16.21 &
  13.29 &
  13.66 &
  14.39 \\
{\color[HTML]{333333} syn-GU\_en-unfiltered+10\%} &
  {\color[HTML]{333333} 66.83} &
  90.11 &
  78.47 &
  17.28 &
  13.27 &
  14.50 &
  15.02 \\
{\color[HTML]{333333} syn-GU\_en-filtered} &
  {\color[HTML]{333333} 67.74} &
  91.35 &
  79.55 &
  \textbf{17.64} &
  \textbf{13.40} &
  14.95 &
  \textbf{15.33} \\
{\color[HTML]{333333} syn-GU\_en-filtered+10\%} &
  {\color[HTML]{333333} \textbf{68.04}} &
  \textbf{92.41} &
  \textbf{80.23} &
  17.62 &
  13.16 &
  \textbf{15.00} &
  15.26 \\ \hline
\end{tabular}
}
}
\caption{Results for Hindi and Gujarati: NLU/NLG tasks on \textit{base-1k} (Hindi) and \textit{mini-1k} (Gujarati) models on different clean and synthetic splits. Test accuracy for NLU tasks; Rouge-L F1 scores for NLG tasks. \textbf{Bold} values represent the best amongst synthetic splits.}
\label{tab:hi-gu-main}
\end{table*}

\section{Results}
\label{sec: results}
We now present our results which help establish the utility of synthetic data for language modeling.

\subsection{Main Results}

In this section, we present results for Hindi, Gujarati, and English language models trained on clean data, as well as synthetic data generated from translations. We demonstrate the impact of filtering and adding additional clean data for extended pretraining of LMs trained solely on synthetic text. Additionally, we observe the effect of using the clean source text along with its translations (synthetic parallel documents) on downstream tasks. We follow the naming convention for different data splits as specified in Section \ref{sec: experiments}. We provide details for the pretraining of each model in Appendix \ref{sec:training hyperparameters appendix}. We provide additional results in Appendix \ref{sec:additional results appendix}.

\noindent\textbf{Filtered Synthetic Data is Competitive with Web Scraped Data:}
The results in Table \ref{tab:hi-gu-main} and \ref{tab:effect of src-en} indicate that \textit{syn-HI\_en-unfiltered}, \textit{syn-GU\_en-unfiltered}, and \textit{syn-EN\_hi-unfiltered} exhibit lower downstream performance compared to their filtered counterparts: \textit{syn-HI\_en-filtered}, \textit{syn-GU\_en-filtered}, and \textit{syn-EN\_hi-filtered}, respectively. It is evident that filtering the synthetic documents using TinyLMs significantly improves the performance of both NLU and NLG tasks.
We also observe that for tasks like CoLA \cite{warstadt-etal-2019-neural}, language models trained solely on synthetic data lag behind when compared to other tasks as seen in Table \ref{tab:en-glue-85M} of Appendix \ref{sec:additional results appendix}. This suggests that \textit{synthetic corpora may lack certain important elements necessary for language models} to perform competitively in linguistic acceptability tasks, as opposed to LMs trained on clean, non-synthetic corpora.
Results for \textit{base-1k} for English are presented in Table \ref{tab:en-glue-85M} in Appendix \ref{sec:additional results appendix} because we focus our attention on Indic languages.

\begin{table*}[htbp!]
\centering
\resizebox{\textwidth}{!}{%
\begin{tabular}{cc|ccccccccc|c}
\hline
\multicolumn{2}{c|}{} &
  {\color[HTML]{333333} \textbf{sst2}} &
  {\color[HTML]{333333} \textbf{cola}} &
  \textbf{mrpc} &
  \textbf{qnli} &
  \textbf{qqp} &
  \textbf{rte} &
  \textbf{mnli-m} &
  \textbf{mnli-mm} &
  \textbf{stsb} &
   \\
\multicolumn{2}{c|}{\multirow{-2}{*}{\textbf{Model}}} &
  {\color[HTML]{333333} acc} &
  {\color[HTML]{333333} mcc} &
  f1 &
  acc &
  f1 &
  acc &
  acc &
  acc &
  pearson &
  \multirow{-2}{*}{\textbf{Avg.}} \\ \hline
\textbf{Original} &
  {\color[HTML]{333333} EN-clean} &
  {\color[HTML]{333333} 87.95} &
  {\color[HTML]{333333} 25.59} &
  83.84 &
  78.83 &
  80.78 &
  64.62 &
  71.6 &
  71.69 &
  73.48 &
  70.93 \\ \hline
 &
  {\color[HTML]{333333} syn-EN\_hi-unfiltered} &
  {\color[HTML]{333333} 87.53} &
  {\color[HTML]{333333} 19.77} &
  79.02 &
  76.49 &
  77.96 &
  55.4 &
  69.65 &
  70.14 &
  67.37 &
  67.04 \\
 &
  {\color[HTML]{333333} syn-EN\_hi-filtered} &
  {\color[HTML]{333333} 87.61} &
  {\color[HTML]{333333} 22.81} &
  81.95 &
  77.63 &
  \textbf{80.57} &
  56.31 &
  70.19 &
  70.89 &
  69.29 &
  68.58 \\
\multirow{-3}{*}{\textbf{\begin{tabular}[c]{@{}c@{}}Translationese\\ Hi$\rightarrow$En\end{tabular}}} &
  {\color[HTML]{333333} syn-EN\_hi-filtered + 10\%} &
  {\color[HTML]{333333} \textbf{87.84}} &
  {\color[HTML]{333333} \textbf{26.61}} &
  \textbf{83.27} &
  \textbf{78.5} &
  80.36 &
  \textbf{61.37} &
  \textbf{71.29} &
  \textbf{71.11} &
  \textbf{71.91} &
  \textbf{70.25} \\ \hline
 &
  {\color[HTML]{333333} syn-EN\_gu-unfiltered} &
  {\color[HTML]{333333} 83.11} &
  {\color[HTML]{333333} 17.66} &
  78.53 &
  66.01 &
  77.68 &
  53.6 &
  63.21 &
  64.55 &
  27.33 &
  59.08 \\
 &
  {\color[HTML]{333333} syn-EN\_gu-filtered} &
  {\color[HTML]{333333} 85.66} &
  {\color[HTML]{333333} 21.15} &
  81.45 &
  66.35 &
  77.36 &
  54.15 &
  66.27 &
  65.72 &
  26.16 &
  60.47 \\
\multirow{-3}{*}{\textbf{\begin{tabular}[c]{@{}c@{}}Translationese\\ Gu$\rightarrow$En\end{tabular}}} &
  {\color[HTML]{333333} syn-EN\_gu-filtered + 10\%} &
  {\color[HTML]{333333} \textbf{86.58}} &
  {\color[HTML]{333333} \textbf{25.17}} &
  \textbf{81.67} &
  \textbf{67.1} &
  \textbf{77.75} &
  \textbf{57.76} &
  \textbf{68.78} &
  \textbf{68.56} &
  \textbf{27.54} &
  \textbf{62.32} \\ \hline
\end{tabular}
}
\caption{Effect of source selection for generating synthetic data on the dev set of GLUE benchmark. All the results reported here are on \textit{mini-1k}.  \textbf{Bold} values represent the best amongst synthetic splits}
\label{tab:effect of src-en}
\end{table*}


\noindent\textbf{Fine-tuning on small amounts of Web Scraped Data boosts performance:}
Even after filtering, we observe that language models trained solely on synthetic text slightly underperform LMs trained on clean text. To address this issue, we conduct extended pretraining of LMs using clean data sourced from IndicMonoDoc. The objective is to determine if this additional training improves performance. We only incorporate an additional 10\% of clean data compared to the LM's previous training data. We see these results across all three languages, and for Hindi and Gujarati, we see that\textit{ by incorporating even a small amount of clean data, we observe an increase in performance on downstream tasks}, bringing the LM at par or closer to the performance of a clean LM. We see an improvement in LMs trained using unfiltered synthetic corpora as well but we believe that filtering leads to the removal of noisy documents and thus better performance. We observe improved performance in language models (LMs) trained with unfiltered synthetic corpora, but filtering out noisy documents enhances performance further. Our ablation study (Table \ref{tab:hi-ablation-85M} in Appendix \ref{sec:additional results appendix}) investigates whether adding 10\% more synthetic data contributes to this improvement or if the data type is key. While performance gains could stem from statistical variances, the consistency across nearly all downstream tasks suggests otherwise.

\begin{figure*}[h!]
    \centering
    \subfigure{\includegraphics[width=0.31\textwidth]{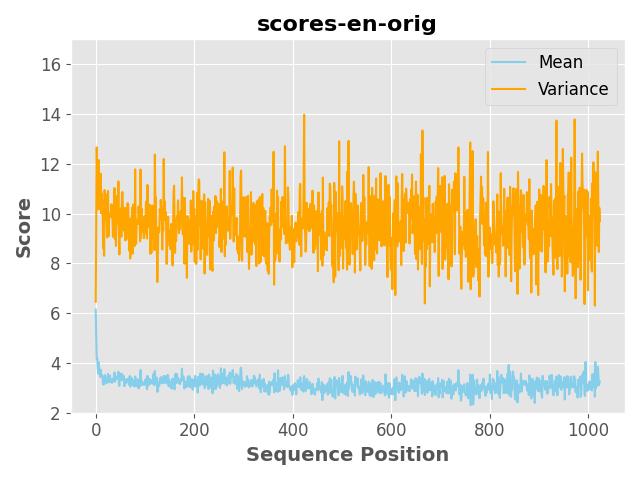}}
    \subfigure{\includegraphics[width=0.31\textwidth]{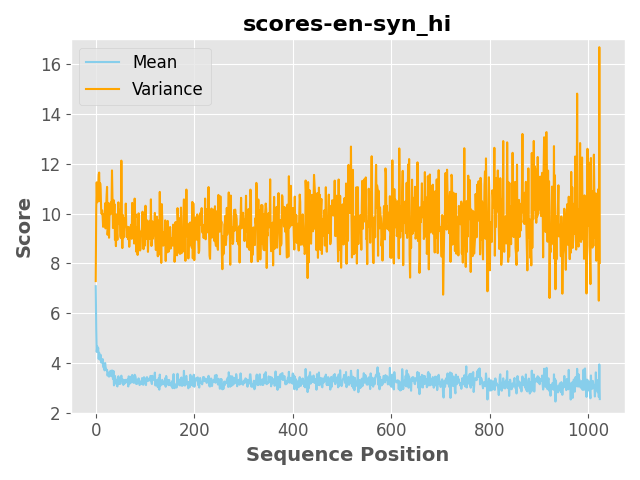}}
    \subfigure{\includegraphics[width=0.31\textwidth]{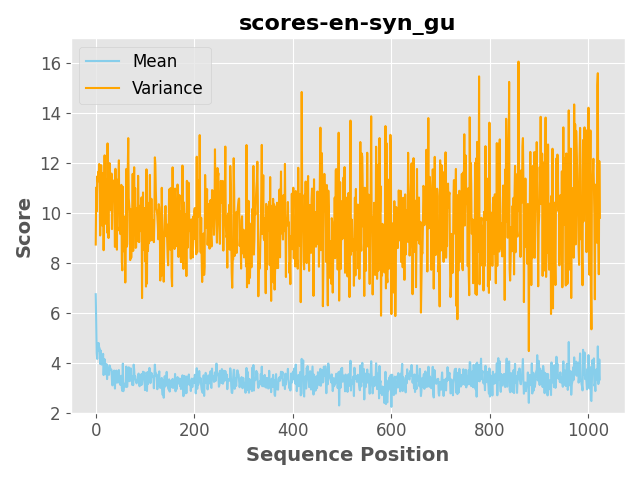}}
    \caption{The plot illustrates TinyLM's perplexity mean and variance across various datasets: Clean-EN (left), Syn-EN from filtered Hindi (middle), and Syn-EN from filtered Gujarati (right). Despite filtering, English documents generated from translating Gujarati show consistently higher variance.}
    \label{fig: variance plot}
\end{figure*}

\noindent\textbf{Impact of source language for synthetic data generation:}
Choosing the right source language for synthetic corpora is crucial, as it influences the characteristics of the generated translationese text. We evaluate this using Hindi and Gujarati clean documents from \textit{IndicMonoDoc}, translating them into English. We use the 1B Indic-En version\footnote{\url{https://huggingface.co/ai4bharat/indictrans2-indic-en-1B}} to translate 5B Hindi tokens and 900M\footnote{Since Gujarati has limited data (900M tokens), we train a \textit{mini-1k} model for fair comparison.} Gujarati tokens into English. In Table \ref{tab:effect of src-en}, we see that the \textit{synthetic text generated from Hindi achieves performance at par with the EN-clean model, while the synthetic text from Gujarati significantly lags behind}. This is likely because Hindi is more \textit{macaronic} than Gujarati, i.e., a lot of Hindi text from the web consists of \textit{Hinglish}, resulting in better translationese text due to increased overlap between languages. This can also be due to the weaker translations generated by the MT model. The performance gap is notable in tasks like STS benchmark, NLI (qnli and mnli), and CoLA, suggesting poorer translation quality from Gu$\rightarrow$ En compared to Hi$\rightarrow$ En.

\begin{table}[htbp!]
\centering
\resizebox{0.98\columnwidth}{!}{%
\begin{tabular}{c|ccccc|c}
\hline
{\color[HTML]{333333} \textbf{Model}} &
  {\color[HTML]{333333} \textbf{iXNLI}} &
  {\color[HTML]{333333} \textbf{bbc-a}} &
  \textbf{iitp-mr} &
  \textbf{iitp-pr} &
  \textbf{midas} &
  \textbf{Avg.} \\ \hline
HI-clean                                     & 68.74                                 & 80.25                        & 67.74 & 77.05 & 78.33 & 74.42 \\ \hline
{\color[HTML]{333333} syn-HI\_en-unfiltered} & {\color[HTML]{333333} 67.32}          & {\color[HTML]{333333} 77.92} & 65.63 & 76.81 & 77.58 & 73.05 \\
{\color[HTML]{333333} syn-HI\_en-filtered}   & {\color[HTML]{333333} 69.48} & {\color[HTML]{333333} 78.98} & 65.16 & 77.43 & 77.33 & 73.68 \\
{\color[HTML]{333333} syn-HI\_en-filtered+10\%} &
  {\color[HTML]{333333} \textbf{70.15}} &
  {\color[HTML]{333333} \textbf{79.56}} &
  \textbf{67.09} &
  \textbf{78.2} &
  \textbf{79.03} &
  \textbf{74.81} \\\hline
\end{tabular}
}
\caption{Effect of reducing model size for Hindi on IndicGLUE accuracy. All the results reported here are on \textit{mini-1k}.  \textbf{Bold} values represent the best amongst synthetic splits}
\label{tab:effect-scaling-hi}
\end{table}


\subsection{Further Exploration}

\noindent\textbf{Analysis of Synthetic Data:}
Figure \ref{fig: variance plot} shows the perplexity mean and variance scores for TinyLM across token positions in the documents. This shows that on unseen documents, TinyLM shows higher variance on English documents generated by translating Gujarati documents from IndicMonoDoc as compared to English clean and English synthetic generated from Hindi. This also gives reason for the deterioration in results in Table \ref{tab:effect of src-en} due to Gujarati documents. Figure \ref{fig: violin plot} shows the distribution of lengths of filtered documents by TinyLMs showing that they do not add any bias for shorter documents during filtering.

\noindent\textbf{Impact of model size:} 
Following Table \ref{tab:effect of src-en} and \ref{tab:effect-scaling-hi}, we see that even after scaling down we see consistent improvements for filtering and adding additional data, which empirically shows that indeed \textit{using synthetic text after filtering is a viable option for pretraining LMs} of varying sizes. In Table \ref{tab:effect-scaling-hi} we see that after filtering and extended pretraining, synthetic text outperforms LMs trained on clean documents from the web in Hindi. This is also supported by our experiments on finetuning Llama-3-8B in Section \ref{sec: scaling}.

\begin{table}[htbp!]
\centering
\resizebox{0.98\columnwidth}{!}{%
\begin{tabular}{c|cccc|c}
\hline
{\color[HTML]{333333} \textbf{Model}} &
  {\color[HTML]{333333} \textbf{\begin{tabular}[c]{@{}c@{}}XLSum\\ HG\end{tabular}}} &
  {\color[HTML]{333333} \textbf{\begin{tabular}[c]{@{}c@{}}XLSum\\ QG\end{tabular}}} &
  \textbf{Cnn} &
  \textbf{Dialogsum} &
  \textbf{Avg.} \\ \hline
{\color[HTML]{333333} EN-clean} &
  23.87 &
  24.05 &
  16.08 &
  20.39 &
  21.10 \\ \hline
{\color[HTML]{333333} syn-EN\_hi-unfiltered} &
  {\color[HTML]{333333} 22.17} &
  {\color[HTML]{333333} 22.97} &
  12.56 &
  18.30 &
  19.00 \\
{\color[HTML]{333333} syn-EN\_hi-filtered} &
  {\color[HTML]{333333} \textbf{23.27}} &
  {\color[HTML]{333333} \textbf{23.83}} &
  \textbf{15.88} &
  \textbf{19.83} &
  \textbf{20.70} \\\hline
\end{tabular}
}
\caption{Performance of English models on NLG tasks. All the results reported here are on \textit{base-1k} and use Rouge-L F1 scores.}
\label{tab:effect nlg}
\end{table}

\noindent\textbf{Impact on NLG:}
Without extended pretraining, language models trained on synthetic text perform as well as those trained on clean documents, suggesting that \textit{for NLG tasks, synthetic data suffices for pretraining}, eliminating the need for clean data. This trend is evident across Hindi, Gujarati, and English NLG results (Tables \ref{tab:hi-gu-main} and \ref{tab:effect nlg}). As their performance matches models trained on clean data, we refrain from extended pretraining for NLG tasks, focusing primarily on abstractive summarization for evaluating generation capabilities.

\subsection{Scaling to Llama-3-8B}
\label{sec: scaling}
To show the effect of using translationese on larger models, we select Llama-3-8B\footnote{\url{https://github.com/meta-llama/llama3}} and Gemma-2B \citep{gemmateam2024gemma} and perform continual pretraining over clean and synthetic data to improve ability over the low resource target language. We take Marathi as a replacement for Hindi for scaling experiments since data for Hindi is abundantly available and existing models already have a good language understanding of Hindi making it harder to compare the effects of utilizing clean vs synthetic data. For a fair comparison, we limit each data split to 344M tokens for Gujarati and 465M tokens for Marathi and follow a similar procedure as described in Section \ref{sec:datafiltering} to generate and filter data for Marathi. We perform extended training for a single epoch using LoRA \citep{DBLP:journals/corr/abs-2106-09685} finetuning on $W_q$, $W_v$ projection matrices using $\alpha$=$16$ and $r$=$8$. We keep the learning rate at $3e^{-5}$ with a weight decay of $0.01$ and an effective batch size of 58k.

\noindent\textbf{Perplexity:} We report average sentence level perplexity on sentences from IN22-Conv and IN22-Gen \cite{gala2023indictrans} in Table \ref{tab:scaling ppl}. We see that filtered synthetic data for Gujarati outperforms clean data, but for Marathi, it does not. This means that \textit{filtering improves performance at scale but relies on the quality of translationese} in the target language. We report perplexities on individual test sets in Appendix \ref{sec:additional results appendix}.

\begin{table}[htbp]
\centering
\resizebox{0.98\columnwidth}{!}{%
\begin{tabular}{ccccc}
\hline
\multicolumn{1}{c|}{\multirow{2}{*}{\textbf{Data}}} &
  \multicolumn{2}{c|}{\textbf{Marathi}} &
  \multicolumn{2}{c}{\textbf{Gujarati}} \\ \cline{2-5} 
\multicolumn{1}{c|}{} &
  \multicolumn{1}{c|}{\textbf{\begin{tabular}[c]{@{}c@{}}Gemma\\ 2B\end{tabular}}} &
  \multicolumn{1}{c|}{\textbf{\begin{tabular}[c]{@{}c@{}}Llama-3\\ 8B\end{tabular}}} &
  \multicolumn{1}{c|}{\textbf{\begin{tabular}[c]{@{}c@{}}Gemma\\ 2B\end{tabular}}} &
  \textbf{\begin{tabular}[c]{@{}c@{}}Llama-3\\ 8B\end{tabular}} \\ \hline
\multicolumn{1}{c|}{Base model} &
  \multicolumn{1}{c|}{178.898} &
  \multicolumn{1}{c|}{66.740} &
  \multicolumn{1}{c|}{71.136} &
  2.839 \\ \hline
\multicolumn{1}{c|}{clean} &
  \multicolumn{1}{c|}{37.599} &
  \multicolumn{1}{c|}{11.196} &
  \multicolumn{1}{c|}{10.350} &
  2.312 \\ \hline
\multicolumn{1}{c|}{synthetic-unfiltered} &
  \multicolumn{1}{c|}{\textbf{92.813}} &
  \multicolumn{1}{c|}{15.697} &
  \multicolumn{1}{c|}{10.941} &
  2.816 \\ 
\multicolumn{1}{c|}{synthetic-filtered} &
  \multicolumn{1}{c|}{104.148} &
  \multicolumn{1}{c|}{\textbf{14.622}} &
  \multicolumn{1}{c|}{\textbf{10.150}} &
  \textbf{2.236} \\ \hline
\end{tabular}
}
\caption{Average perplexity ($\downarrow$) of models trained on Translationese vs. Clean data on IN22-Conv and IN22-Gen test sets shows improvement with large-scale models. \textbf{Bold} represents best among synthetic data splits.}
\label{tab:scaling ppl}
\end{table}

\noindent\textbf{Few Shot Prompting:} We evaluate our continually pre-trained models using few-shot prompting on IndicSentiment classification \cite{doddapaneni2023towards} as the NLU task and En$\rightarrow$Indic machine translation on IN22-Gen and FloRes-200 as the NLG task. Prompts used are shown in Appendix \ref{app: prompts} with examples are randomly sampled from the validation set for FloRes and other examples from the IN22-Gen test set, ensuring no example is repeated in the prompts. We use a beam width of 5 with early stopping enabled. Results are shown in Tables \ref{tab: scaling MT} and \ref{tab: scaling IndicSentiment}. We see that filtering improves performance on MT when compared to synthetic splits, but IndicSentiment, shows only marginal improvements. Nonetheless, models trained on filtered data show lower perplexity and better performance in few-shot settings, indicating their promise. We leave the exploration of training multilingual LLMs on large-scale translationese data for future research.

\begin{table}[]
\centering
\resizebox{0.98\columnwidth}{!}{%
\begin{tabular}{ccccc}
\hline
\multicolumn{1}{c|}{\multirow{2}{*}{\textbf{Data}}} & \multicolumn{2}{c|}{\textbf{Flores-200}}                & \multicolumn{2}{c}{\textbf{IN22-Gen}} \\ \cline{2-5} 
\multicolumn{1}{c|}{} &
  \multicolumn{1}{c|}{\textbf{Marathi}} &
  \multicolumn{1}{c|}{\textbf{Gujarati}} &
  \multicolumn{1}{c|}{\textbf{Marathi}} &
  \textbf{Gujarati} \\ \hline
\multicolumn{1}{c|}{Base model}                     & \multicolumn{1}{c|}{27.83} & \multicolumn{1}{c|}{34.35} & \multicolumn{1}{c|}{30}      & 34.92  \\ \hline
\multicolumn{1}{c|}{clean}                          & \multicolumn{1}{c|}{34.02} & \multicolumn{1}{c|}{35.61} & \multicolumn{1}{c|}{33.94}   & 35.42  \\ \hline
\multicolumn{1}{c|}{synthetic-unfiltered}           & \multicolumn{1}{c|}{30.63} & \multicolumn{1}{c|}{34.19} & \multicolumn{1}{c|}{29.67}   & 32.1   \\ 
\multicolumn{1}{c|}{synthetic-filtered} &
  \multicolumn{1}{c|}{\textbf{31.81}} &
  \multicolumn{1}{c|}{\textbf{35.54}} &
  \multicolumn{1}{c|}{\textbf{31.3}} &
  \textbf{35} \\ \hline
\end{tabular}
}
\caption{chrF++ scores on 5-shot Machine Translation FloRes and IN22-Gen test sets on Llama-3-8B. \textbf{Bold} represents best among synthetic data experiments.}
\label{tab: scaling MT}
\end{table}

\begin{table}[]
\centering
\resizebox{0.98\columnwidth}{!}{%
\begin{tabular}{ccccc}
\hline
\multicolumn{1}{c|}{\multirow{2}{*}{\textbf{Data}}} &
  \multicolumn{2}{c|}{\textbf{Marathi}} &
  \multicolumn{2}{c}{\textbf{Gujarati}} \\ \cline{2-5} 
\multicolumn{1}{c|}{} &
  \multicolumn{1}{c|}{\textbf{\begin{tabular}[c]{@{}c@{}}Gemma\\ 2B\end{tabular}}} &
  \multicolumn{1}{c|}{\textbf{\begin{tabular}[c]{@{}c@{}}Llama-3\\ 8B\end{tabular}}} &
  \multicolumn{1}{c|}{\textbf{\begin{tabular}[c]{@{}c@{}}Gemma\\ 2B\end{tabular}}} &
  \textbf{\begin{tabular}[c]{@{}c@{}}Llama-3\\ 8B\end{tabular}} \\ \hline
\multicolumn{1}{c|}{Base model} &
  \multicolumn{1}{c|}{90.89 $^{\pm 0.005}$} &
  \multicolumn{1}{c|}{95 $^{\pm 0.009}$} &
  \multicolumn{1}{c|}{83.84 $^{\pm 0.002}$} &
  92.69 $^{\pm 0.018}$ \\ \hline
\multicolumn{1}{c|}{clean} &
  \multicolumn{1}{c|}{90 $^{\pm 0.014}$} &
  \multicolumn{1}{c|}{97.17 $^{\pm 0.013}$} &
  \multicolumn{1}{c|}{87.79 $^{\pm 0.012}$} &
  93.33 $^{\pm 0.0132}$ \\ \hline
\multicolumn{1}{c|}{\begin{tabular}[c]{@{}c@{}}synthetic\\ unfiltered\end{tabular}} &
  \multicolumn{1}{c|}{\textbf{89.66 $^{\pm 0.012}$}} &
  \multicolumn{1}{c|}{95.38 $^{\pm 0.016}$} &
  \multicolumn{1}{c|}{83.88 $^{\pm 0.009}$} &
  92.81 $^{\pm 0.01 }$\\ \hline
\multicolumn{1}{c|}{\begin{tabular}[c]{@{}c@{}}synthetic\\ filtered\end{tabular}} &
  \multicolumn{1}{c|}{86.67 $^{\pm 0.016}$} &
  \multicolumn{1}{c|}{\textbf{96.15 $^{\pm 0.011}$}} &
  \multicolumn{1}{c|}{\textbf{84.10 $^{\pm 0.013}$}} &
  \textbf{92.94 $^{\pm 0.007}$} \\ \hline
\end{tabular}
}
\caption{Average accuracy with standard deviation (superscript) over 5 runs on 10-shot IndicSentiment classification task. \textbf{Bold} represents the best among synthetic data experiments.}
\label{tab: scaling IndicSentiment}
\end{table}

\section{Conclusion}
\label{sec: conclusion}

In this paper, we performed a first of its kind study showing the promise of using translationese data for training language models for low-resource languages. Our simple pipeline involves the translation of high-resource source language documents at scale, followed by perplexity based filtering using small and efficient language models trained on clean target low-resource language data. We then showed on a variety of downstream natural language understanding and generative tasks that both small and large language models pre-trained on clean synthetic data are comparable to those trained on clean data. While we observed that the source language for synthetic data generation matters, it is clear that synthetic data can help bridge the resource scarcity faced by a vast majority of languages for language modeling. Future work will focus on better and faster synthetic data generation and filtering mechanisms.

\section*{Acknowledgements}
We extend our thanks to the Department of Computer Science and Engineering at the Indian Institute of Technology (IIT) Bombay for providing access to GPU servers, and to the Centre for Development of Advanced Computing (C-DAC) for granting us access to the Param Siddhi Supercomputer. These computational resources were essential for the successful completion of this work.

\section*{Limitations}
We consider the following limitations of our work.

\begin{itemize}
    \item We show that synthetic data also helps for larger models like Llama-3-8B but for even larger models above 100B parameters, effects of translationese may be different. However, synthetic data generated from translations can surely help fill knowledge gaps.
    \item Due to the extensive size of the test sets for IndicNLG tasks (Question Generation, WikiBio generation, Headline Generation, and Sentence Summarization), we couldn't experiment with them in their entirety. However, since we already use 4000 examples per language, we anticipate that the overall trends remain unchanged.
    \item We report GLUE validation set results for all models due to the large scale of our experiments, following existing practices. Our goal is to demonstrate synthetic data utility, not to achieve state-of-the-art results.
    \item Our framework heavily relies on the translation model's performance. Despite this dependency, we are confident that our approach will significantly enhance the performance of mid-resource languages, especially where the translation model is already proficient.
\end{itemize}

\section*{Ethical Considerations}
As a part of this paper, we release monolingual and synthetic data. While we have taken care to remove any toxic content, accidental occurrences may exist and thus we exercise caution when using our data for training language models as they may produce toxic outputs. Given that we have shown the utility of synthetic data for training LMs, it should be possible to mass produce synthetic toxic data in various languages leading to LMs that can generate multilingual toxic content. However, this opens up research opportunities on how to detect and filter toxic content from synthetically created data.

We release the code and models with an MIT License\footnote{\url{https://opensource.org/license/mit/}}. The dataset is released under a CC-0 License\footnote{\url{https://creativecommons.org/share-your-work/public-domain/cc0/}}.

\bibliography{anthology,custom}

\appendix

\section{Additional results}
\label{sec:additional results appendix}
We report additional results in this section. Tables \ref{tab:translation chrf appendix}, \ref{tab:translation bleu appendix} show the chrF++ and BLEU scores across three translation evaluation benchmarks. This shows that using parallel synthetic data does not deteriorate the performance of the language model. Similar results are shown in Table \ref{tab:nlg-hi bilingual appendix} for IndicNLG tasks where performance on Hindi generation tasks are only affected by a small margin and coupled with results in Table \ref{tab:en-glue-85M} showing that scores are not affected by using Hindi synthetic parallel data.

\begin{table*}[]
\centering
\resizebox{\textwidth}{!}{%
\begin{tabular}{cc|ccccccccc|c}
\hline
\multicolumn{2}{c|}{} &
  {\color[HTML]{333333} \textbf{sst2}} &
  {\color[HTML]{333333} \textbf{cola}} &
  \textbf{mrpc} &
  \textbf{qnli} &
  \textbf{qqp} &
  \textbf{rte} &
  \textbf{mnli-m} &
  \textbf{mnli-mm} &
  \textbf{stsb} &
   \\
\multicolumn{2}{c|}{\multirow{-2}{*}{\textbf{Model}}} &
  {\color[HTML]{333333} acc} &
  {\color[HTML]{333333} mcc} &
  f1 &
  acc &
  f1 &
  acc &
  acc &
  acc &
  pearson &
  \multirow{-2}{*}{\textbf{Avg.}} \\ \hline
 &
  {\color[HTML]{333333} EN-clean} &
  {\color[HTML]{333333} 90.94} &
  {\color[HTML]{333333} 40.26} &
  87.4 &
  84.98 &
  84.47 &
  65.34 &
  77.84 &
  77.96 &
  82.67 &
  76.87 \\ \cline{2-12} 
 &
  {\color[HTML]{333333} syn-EN\_hi-unfiltered} &
  {\color[HTML]{333333} 84.61} &
  {\color[HTML]{333333} 31.1} &
  81.78 &
  79.35 &
  81.44 &
  63.3 &
  72.94 &
  73.16 &
  78.9 &
  71.84 \\
 &
  {\color[HTML]{333333} syn-EN\_hi-unfiltered + 10\%} &
  {\color[HTML]{333333} 87.39} &
  {\color[HTML]{333333} 34.22} &
  85.77 &
  80.96 &
  81.07 &
  65.11 &
  74.76 &
  74.38 &
  80.32 &
  73.78 \\
 &
  {\color[HTML]{333333} syn-EN\_hi-filtered} &
  {\color[HTML]{333333} 88.3} &
  {\color[HTML]{333333} 34.03} &
  \textbf{86.55} &
  83.59 &
  83.64 &
  63.17 &
  75.6 &
  75.41 &
  81.1 &
  74.60 \\ &
  {\color[HTML]{333333} syn-EN\_hi-filtered + 10\%} &
  {\color[HTML]{333333} \textbf{90.13}} &
  {\color[HTML]{333333} \textbf{35.75}} &
  86.41 &
  \textbf{84.75} &
  \textbf{84.21} &
  \textbf{65.34} &
  \textbf{76.99} &
  \textbf{76.91} &
  \textbf{81.95} &
  \textbf{75.83} \\ \hline
\end{tabular}
}
\caption{Results on English: Dev set of GLUE tasks for different synthetic splits on the \textit{base-1k} model. Synthetic LMs perform almost as well as clean LMs after filtering and further training with clean data.  \textbf{Bold} values represent the best amongst synthetic splits.}
\label{tab:en-glue-85M}
\end{table*}

\noindent\textbf{Using synthetic for one language doesn't impact performance in another:}
For many multilingual language models, data imbalance causes a gap in performance across languages. But what if we can combine synthetic data along with clean data for training multilingual models? would the synthetic part deteriorate the performance of the multilingual model? To experiment with this, we train bilingual \textit{base-1k} models over different combinations of clean and synthetic corpora for English and Hindi and evaluate their performance on GLUE \cite{wang2018glue}, and report performance on IndicNLG, and Machine translation in Appendix \ref{sec:additional results appendix}. Following Table \ref{tab:bi-en-glue-85M}, we see that using Hindi synthetic data does not affect its performance compared to \textit{BI-EN-HI-clean} model which is solely trained on clean corpora. This implies that \textit{it is possible to train multilingual models where some languages are trained only over a clean subset and others on synthetic} without deteriorating performance across languages. We further see that \textit{using parallel data does not have much impact on multilingual models}.

\begin{table*}[]
\centering
\resizebox{\textwidth}{!}{%
\begin{tabular}{cc|ccccccccc|c}
\hline
\multicolumn{2}{c|}{} &
  {\color[HTML]{333333} \textbf{sst2}} &
  {\color[HTML]{333333} \textbf{cola}} &
  \textbf{mrpc} &
  \textbf{qnli} &
  \textbf{qqp} &
  \textbf{rte} &
  \textbf{mnli-m} &
  \textbf{mnli-mm} &
  \textbf{stsb} &
   \\
\multicolumn{2}{c|}{\multirow{-2}{*}{\textbf{Model}}} &
  {\color[HTML]{333333} acc} &
  {\color[HTML]{333333} mcc} &
  f1 &
  acc &
  f1 &
  acc &
  acc &
  acc &
  pearson &
  \multirow{-2}{*}{\textbf{Avg.}} \\ \hline
 &
  {\color[HTML]{333333} BI-EN-HI-clean} &
  {\color[HTML]{333333} 89.56} &
  {\color[HTML]{333333} 38.53} &
  85.56 &
  84.88 &
  84.39 &
  64.25 &
  76.4 &
  77.27 &
  82.07 &
  75.88 \\ \cline{2-12} 
 &
  {\color[HTML]{333333} BI-EN-HI\_syn-parallel-filtered} &
  {\color[HTML]{333333} 89.56} &
  {\color[HTML]{333333} \textbf{39.57}} &
  85.71 &
  84.75 &
  \textbf{84.62} &
  64.98 &
  \textbf{77.31} &
  \textbf{77.85} &
  82.41 &
  76.31 \\
 &
  {\color[HTML]{333333} BI-EN-HI\_syn-nonparallel-filtered} &
  {\color[HTML]{333333} \textbf{89.79}} &
  {\color[HTML]{333333} 38.68} &
  \textbf{86.92} &
  \textbf{85.08} &
  84.06 &
  65.34 &
  77.15 &
  77.55 &
  \textbf{83.01} &
  \textbf{76.40} \\
 &
  {\color[HTML]{333333} BI-EN\_syn-HI\_syn-filtered} &
  {\color[HTML]{333333} 87.95} &
  {\color[HTML]{333333} 30.05} &
  84.9 &
  83.7 &
  83.97 &
  63.89 &
  75.63 &
  76.24 &
  82.24 &
  74.29 \\ &
  {\color[HTML]{333333} BI-EN\_syn-HI\_syn-filtered + 10\%} &
  {\color[HTML]{333333} 89.1} &
  {\color[HTML]{333333} 35.45} &
  85.34 &
  84.53 &
  84.18 &
  \textbf{65.7} &
  76.64 &
  77.24 &
  82.1 &
  75.59 \\\hline
\end{tabular}
}
\caption{Results on English for Bilingual models: Dev set of GLUE tasks for different synthetic splits on the \textit{base-1k} model. Training bilingual models using synthetic data in one language (Hindi) does not affect the performance in the other language (English).  \textbf{Bold} values represent the best amongst synthetic splits.}
\label{tab:bi-en-glue-85M}
\end{table*}

\begin{table}[htbp!]
\centering
\resizebox{0.96\columnwidth}{!}{%
\begin{tabular}{c|ccc}
\hline
{\color[HTML]{333333} }                                & \multicolumn{3}{c}{\textbf{FLORES}}                                  \\ \cline{2-4} 
\multirow{-2}{*}{{\color[HTML]{333333} \textbf{Model}}}   & \textbf{EN-HI} & \multicolumn{1}{c|}{\textbf{HI-EN}} & \textbf{Avg.}  \\ \hline
{\color[HTML]{333333} BI-EN-HI-clean}                  & 46.56          & \multicolumn{1}{c|}{51.7}           & 49.13         \\ \hline
{\color[HTML]{333333} BI-EN-HI\_syn-parallel-filtered} & 44.12          & \multicolumn{1}{c|}{50.64}          & 47.38         \\
{\color[HTML]{333333} BI-EN-HI\_syn-nonparallel-filtered} & \textbf{45.65} & \multicolumn{1}{c|}{\textbf{51.29}} & \textbf{48.47} \\ \hline
{\color[HTML]{333333} }                                & \textbf{EN-GU} & \multicolumn{1}{c|}{\textbf{GU-EN}} & \textbf{Avg.} \\ \hline
{\color[HTML]{333333} BI-EN-GU-clean}                  & 26.44          & \multicolumn{1}{c|}{35.3}           & 30.87         \\ \hline
{\color[HTML]{333333} BI-EN-GU\_syn-parallel-filtered}    & \textbf{26.77} & \multicolumn{1}{c|}{34.84}          & 30.81          \\
{\color[HTML]{333333} BI-EN-GU\_syn-nonparallel-filtered} & 26.7           & \multicolumn{1}{c|}{\textbf{36.54}} & \textbf{31.62} \\\hline
\end{tabular}
}
\caption{chrF++ Scores on FLoRes translation task. EN-HI models are based on \textit{base-1k} and EN-GU models are based on \textit{mini-1k}}
\label{tab:effect translation}
\end{table}

\noindent\textbf{Impact on Machine Translation: (MT)} 
We focus on MT separately as a special case of NLG. We hypothesized that using parallel synthetic documents for bilingual models would improve translation performance by enhancing alignment between languages. However, our evaluation fails this hypothesis. Results indicate that \textit{using nonparallel synthetic documents yields similar translation performance across language directions and benchmarks compared to parallel synthetic documents}. This might be because there is no explicit alignment happening during training between parallel documents. See Table \ref{tab:effect translation} for chrF++ scores on FLoRes-200 \cite{nllbteam2022language}, and Appendix \ref{sec:additional results appendix} for chrF++ and BLEU scores on IN22-Conv, IN22-Gen \cite{gala2023indictrans}.

\begin{table}[]
\centering
\resizebox{0.99\columnwidth}{!}{%
\begin{tabular}{ccccc}
\hline
\multicolumn{5}{c}{\textbf{Perplexity-IN22 Conv}} \\ \hline
\multicolumn{1}{c|}{\multirow{2}{*}{\textbf{Data}}} &
  \multicolumn{2}{c|}{\textbf{Marathi}} &
  \multicolumn{2}{c}{\textbf{Gujarati}} \\ \cline{2-5} 
\multicolumn{1}{c|}{} &
  \multicolumn{1}{c|}{\textbf{Gemma-2B}} &
  \multicolumn{1}{c|}{\textbf{Llama-3-8B}} &
  \multicolumn{1}{c|}{\textbf{Gemma-2B}} &
  \textbf{Llama-3-8B} \\ \hline
\multicolumn{1}{c|}{Base model} &
  \multicolumn{1}{c|}{332.9036} &
  \multicolumn{1}{c|}{121.6586} &
  \multicolumn{1}{c|}{131.8027} &
  3.3922 \\ \hline
\multicolumn{1}{c|}{clean} &
  \multicolumn{1}{c|}{58.3804} &
  \multicolumn{1}{c|}{15.1342} &
  \multicolumn{1}{c|}{12.3128} &
  2.5208 \\ \hline
\multicolumn{1}{c|}{synthetic-unfiltered} &
  \multicolumn{1}{c|}{\textbf{168.0933}} &
  \multicolumn{1}{c|}{23.018} &
  \multicolumn{1}{c|}{12.7995} &
  3.0671 \\ 
\multicolumn{1}{c|}{synthetic-filtered} &
  \multicolumn{1}{c|}{191.1506} &
  \multicolumn{1}{c|}{\textbf{21.355}} &
  \multicolumn{1}{c|}{\textbf{11.6916}} &
  \textbf{2.3849} \\ \hline
\end{tabular}}
\caption{Perplexity on IN-22 Conv. Bold values represent best among synthetic splits.}
\label{tab:my-table}
\end{table}

\begin{table}[]
\centering
\resizebox{0.99\columnwidth}{!}{%
\begin{tabular}{ccccc}
\hline
\multicolumn{5}{c}{\textbf{Perplexity-IN22 Gen}}                                                                                                         \\ \hline
\multicolumn{1}{c|}{\multirow{2}{*}{\textbf{Data}}} & \multicolumn{2}{c|}{\textbf{Marathi}}                      & \multicolumn{2}{c}{\textbf{Gujarati}} \\ \cline{2-5} 
\multicolumn{1}{c|}{} &
  \multicolumn{1}{c|}{\textbf{Gemma-2B}} &
  \multicolumn{1}{c|}{\textbf{Llama-3-8B}} &
  \multicolumn{1}{c|}{\textbf{Gemma-2B}} &
  \textbf{Llama-3-8B} \\ \hline
\multicolumn{1}{c|}{Base model}                     & \multicolumn{1}{c|}{24.893}  & \multicolumn{1}{c|}{11.821} & \multicolumn{1}{c|}{10.4693} & 2.2853 \\ \hline
\multicolumn{1}{c|}{clean}                          & \multicolumn{1}{c|}{16.8172} & \multicolumn{1}{c|}{7.2572} & \multicolumn{1}{c|}{8.3868}  & 2.1032 \\ \hline
\multicolumn{1}{c|}{synthetic-unfiltered} &
  \multicolumn{1}{c|}{\textbf{17.5333}} &
  \multicolumn{1}{c|}{8.3766} &
  \multicolumn{1}{c|}{9.0821} &
  2.564 \\ 
\multicolumn{1}{c|}{synthetic-filtered} &
  \multicolumn{1}{c|}{17.1451} &
  \multicolumn{1}{c|}{\textbf{7.8885}} &
  \multicolumn{1}{c|}{\textbf{8.6089}} &
  \textbf{2.0868} \\ \hline
\end{tabular}}
\caption{Perplexity on IN-22 Gen. Bold values represent best among synthetic splits.}
\label{tab:my-table}
\end{table}

\begin{table*}[]
\centering
\resizebox{\textwidth}{!}{%
\begin{tabular}{c|cccccc|ccccc}
\hline
{\color[HTML]{333333} } &
  \multicolumn{6}{c|}{{\color[HTML]{333333} \textbf{NLU}}} &
  \multicolumn{5}{c}{\textbf{NLG}} \\ \cline{2-12} 
\multirow{-2}{*}{{\color[HTML]{333333} \textbf{Model}}} &
  {\color[HTML]{333333} \textbf{iXNLI}} &
  \textbf{bbc-a} &
  \textbf{iitp-mr} &
  \textbf{iitp-pr} &
  \textbf{midas} &
  \multicolumn{1}{l|}{\textbf{Avg.}} &
  \textbf{\begin{tabular}[c]{@{}c@{}}Headline\\ Gen.\end{tabular}} &
  \textbf{\begin{tabular}[c]{@{}c@{}}Sentence\\ Summ.\end{tabular}} &
  \textbf{\begin{tabular}[c]{@{}c@{}}Question\\ Gen.\end{tabular}} &
  \textbf{Wikibio} &
  \multicolumn{1}{l}{\textbf{Avg.}} \\ \hline
{\color[HTML]{333333} HI-clean} &
  {\color[HTML]{333333} 73.61} &
  81.75 &
  72.58 &
  79.73 &
  80.34 &
  77.60 &
  27.54 &
  23.64 &
  24.84 &
  52.16 &
  32.04 \\ \hline
{\color[HTML]{333333} syn-HI\_en-unfiltered} &
  {\color[HTML]{333333} 72.87} &
  77.92 &
  64.36 &
  76.22 &
  79.91 &
  74.26 &
  27.29 &
  22.93 &
  24.22 &
  50.14 &
  31.14 \\
{\color[HTML]{333333} syn-HI\_en-unfiltered+10\% clean} &
  {\color[HTML]{333333} 74.63} &
  78.36 &
  67.75 &
  77.46 &
  80.17 &
  75.67 &
  26.98 &
  23.20 &
  24.76 &
  \textbf{51.34} &
  31.57 \\
{\color[HTML]{333333} syn-HI\_en-filtered} &
  {\color[HTML]{333333} \textbf{74.75}} &
  \textbf{81.06} &
  69.03 &
  78.58 &
  79.73 &
  76.63 &
  27.15 &
  23.10 &
  24.41 &
  49.88 &
  31.13 \\
{\color[HTML]{333333} syn-HI\_en-filtered+10\% clean} &
  {\color[HTML]{333333} 74.49} &
  80.94 &
  71.61 &
  \textbf{79.92} &
  \textbf{80.64} &
  \textbf{77.52} &
  \textbf{27.87} &
  24.23 &
  \textbf{24.87} &
  51.18 &
  \textbf{32.04} \\
{\color[HTML]{333333} syn-HI\_en-filtered+10\% synthetic} &
  {\color[HTML]{333333} 74.95} &
  80.83 &
  \textbf{71.93} &
  \textbf{79.92} &
  78.73 &
  77.27 &
  27.64 &
  \textbf{25.13} &
  23.60 &
  43.48 &
  29.96 \\  \hline
\end{tabular}
} 					
\caption{Ablation Results for Hindi using additional 10\% clean vs synthetic data: NLU/NLG tasks on \textit{base-1k} (Hindi) different clean and synthetic splits. Test accuracy for NLU tasks; Rouge-L F1 scores for NLG tasks. \textbf{Bold} values represent the best amongst synthetic splits.}
\label{tab:hi-ablation-85M}
\end{table*}

\begin{table*}[htbp!]
\centering
\resizebox{1.95\columnwidth}{!}{%
\begin{tabular}{c|cccccc}
\hline
{\color[HTML]{333333} } &
  \multicolumn{2}{c}{{\color[HTML]{333333} \textbf{IN22-Conv}}} &
  \multicolumn{2}{c}{\textbf{IN22-Gen}} &
  \multicolumn{2}{c}{\textbf{FLORES}} \\ \cline{2-7} 
\multirow{-2}{*}{{\color[HTML]{333333} \textbf{Model}}} &
  {\color[HTML]{333333} EN-HI} &
  {\color[HTML]{333333} HI-EN} &
  EN-HI &
  HI-EN &
  EN-HI &
  HI-EN \\ \hline
{\color[HTML]{333333} BI-EN-HI-clean} &
  {\color[HTML]{333333} 41.22} &
  {\color[HTML]{333333} 50.3} &
  43.49 &
  47.83 &
  46.56 &
  51.7 \\ \hline
{\color[HTML]{333333} BI-EN-HI\_syn-parallel-filtered} &
  {\color[HTML]{333333} \textbf{41.92}} &
  {\color[HTML]{333333} \textbf{49.67}} &
  41.61 &
  46.95 &
  44.12 &
  50.64 \\
{\color[HTML]{333333} BI-EN-HI\_syn-nonparallel-filtered} &
  {\color[HTML]{333333} 40.74} &
  {\color[HTML]{333333} 49.54} &
  \textbf{42.28} &
  \textbf{47.66} &
  \textbf{45.65} &
  \textbf{51.29} \\ \hline
\multicolumn{1}{l|}{{\color[HTML]{333333} }} &
  {\color[HTML]{333333} EN-GU} &
  {\color[HTML]{333333} GU-EN} &
  EN-GU &
  GU-EN &
  EN-GU &
  GU-EN \\ \hline
{\color[HTML]{333333} BI-EN-GU-clean} &
  {\color[HTML]{333333} 35.85} &
  {\color[HTML]{333333} 41.27} &
  22.95 &
  31.83 &
  26.44 &
  35.3 \\ \hline
{\color[HTML]{333333} BI-EN-GU\_syn-parallel-filtered} &
  34.36 &
  41.86 &
  22.93 &
  30.84 &
  \textbf{26.77} &
  34.84 \\
{\color[HTML]{333333} BI-EN-GU\_syn-nonparallel-filtered} &
  \textbf{34.49} &
  \textbf{42.08} &
  \textbf{23.06} &
  \textbf{32.81} &
  26.7 &
  \textbf{36.54}
\end{tabular}
}
\caption{chrF++ Scores on FloRes, IN22-Conv and IN22-Gen splits for translation task. EN-HI models are based on \textit{base-1k} and EN-GU models are based on \textit{mini-1k}. \textbf{Bold} values
represent the best amongst synthetic splits.}
\label{tab:translation chrf appendix}
\end{table*}

\begin{table*}[htbp!]
\centering
\resizebox{1.95\columnwidth}{!}{%
\begin{tabular}{c|cccccc}
\hline
{\color[HTML]{333333} } &
  \multicolumn{2}{c}{{\color[HTML]{333333} \textbf{IN22-Conv}}} &
  \multicolumn{2}{c}{\textbf{IN22-Gen}} &
  \multicolumn{2}{c}{\textbf{FLORES}} \\ \cline{2-7} 
\multirow{-2}{*}{{\color[HTML]{333333} \textbf{Model}}} &
  {\color[HTML]{333333} EN-HI} &
  {\color[HTML]{333333} HI-EN} &
  EN-HI &
  HI-EN &
  EN-HI &
  HI-EN \\ \hline
{\color[HTML]{333333} BI-EN-HI-clean} &
  {\color[HTML]{333333} 19.58} &
  {\color[HTML]{333333} 23.01} &
  17.23 &
  19.72 &
  21.8 &
  21.73 \\ \hline
{\color[HTML]{333333} BI-EN-HI\_syn-parallel-filtered} &
  {\color[HTML]{333333} \textbf{19.64}} &
  {\color[HTML]{333333} \textbf{23.79}} &
  \textbf{16.57} &
  \textbf{20.14} &
  \textbf{21.63} &
  \textbf{22.6} \\
{\color[HTML]{333333} BI-EN-HI\_syn-nonparallel-filtered} &
  {\color[HTML]{333333} 19.25} &
  {\color[HTML]{333333} 22.47} &
  16.37 &
  19.74 &
  21.51 &
  21.74 \\ \hline
\multicolumn{1}{l|}{{\color[HTML]{333333} }} &
  {\color[HTML]{333333} EN-GU} &
  {\color[HTML]{333333} GU-EN} &
  EN-GU &
  GU-EN &
  EN-GU &
  GU-EN \\ \hline
{\color[HTML]{333333} BI-EN-GU-clean} &
  {\color[HTML]{333333} 10.24} &
  {\color[HTML]{333333} 15.19} &
  4.65 &
  7.92 &
  5.44 &
  9.57 \\ \hline
{\color[HTML]{333333} BI-EN-GU\_syn-parallel-filtered} &
  \textbf{11.24} &
  \textbf{15.7} &
  4.87 &
  8.44 &
  \textbf{6.7} &
  10.02 \\
{\color[HTML]{333333} BI-EN-GU\_syn-nonparallel-filtered} &
  10.86 &
  15.57 &
  \textbf{5.07} &
  \textbf{9.07} &
  6.17 &
  \textbf{10.03}
\end{tabular}
}
\caption{BLEU Scores on FloRes, IN22-Conv and IN22-Gen splits for translation task. EN-HI models are based on \textit{base-1k} and EN-GU models are based on \textit{mini-1k}. \textbf{Bold} values
represent the best amongst synthetic splits.}
\label{tab:translation bleu appendix}
\end{table*}

\begin{table*}[htbp!]
\centering
\resizebox{1.95\columnwidth}{!}{%
\begin{tabular}{c|cccc}
\hline
{\color[HTML]{333333} \textbf{Model}} &
  {\color[HTML]{333333} \textbf{\begin{tabular}[c]{@{}c@{}}Headline\\ Generation\end{tabular}}} &
  {\color[HTML]{333333} \textbf{\begin{tabular}[c]{@{}c@{}}Sentence\\ Summarization\end{tabular}}} &
  \textbf{\begin{tabular}[c]{@{}c@{}}Question\\ Generation\end{tabular}} &
  \textbf{\begin{tabular}[c]{@{}c@{}}Wikibio\\ Generation\end{tabular}} \\ \hline
{\color[HTML]{333333} BI-EN-HI-clean} &
  {\color[HTML]{333333} 27.47} &
  {\color[HTML]{333333} 23.78} &
  24.25 &
  50.82 \\ \hline
{\color[HTML]{333333} BI-EN-HI\_syn-parallel-filtered} &
  {\color[HTML]{333333} 26.96} &
  {\color[HTML]{333333} \textbf{23.10}} &
  \textbf{25.38} &
  48.26 \\
{\color[HTML]{333333} BI-EN-HI\_syn-nonparallel-filtered} &
  {\color[HTML]{333333} \textbf{27.32}} &
  {\color[HTML]{333333} 22.84} &
  24.95 &
  \textbf{50.22}
\end{tabular}
}
\caption{Performance of Bilingual models on IndicNLG tasks. All the results reported here are on base-1k and use Rouge-L F1 scores. \textbf{Bold} values represent the best amongst synthetic splits.}
\label{tab:nlg-hi bilingual appendix}
\end{table*}

\begin{table}[htbp]
\centering
\resizebox{0.98\columnwidth}{!}{%
\begin{tabular}{ccccc}
\hline
\multicolumn{5}{c}{\textbf{IndicXNLI 10-Shot}} \\ \hline
\multicolumn{1}{c|}{\multirow{2}{*}{\textbf{Data}}} &
  \multicolumn{2}{c|}{\textbf{Marathi}} &
  \multicolumn{2}{c}{\textbf{Gujarati}} \\ \cline{2-5} 
\multicolumn{1}{c|}{} &
  \multicolumn{1}{c|}{\textbf{Gemma-2B}} &
  \multicolumn{1}{c|}{\textbf{Llama-3-8B}} &
  \multicolumn{1}{c|}{\textbf{Gemma-2B}} &
  \textbf{Llama-3-8B} \\ \hline
\multicolumn{1}{c|}{Base model} &
  \multicolumn{1}{c|}{\textbf{36.56 ± 0.47}} &
  \multicolumn{1}{c|}{50.66 ± 1.46} &
  \multicolumn{1}{c|}{\textbf{34.74 ± 0.83}} &
  47.63 ± 1.96 \\ \hline
\multicolumn{1}{c|}{clean} &
  \multicolumn{1}{c|}{35.92 ± 0.79} &
  \multicolumn{1}{c|}{48.47 ± 1.44} &
  \multicolumn{1}{c|}{33.40 ± 0.29} &
  48.01 ± 0.99 \\ \hline
\multicolumn{1}{c|}{synthetic-unfiltered} &
  \multicolumn{1}{c|}{35.66 ± 0.93} &
  \multicolumn{1}{c|}{49.50 ± 1.76} &
  \multicolumn{1}{c|}{33.26 ± 0.39} &
  \textbf{48.22 ± 1.88} \\ \hline
\multicolumn{1}{c|}{synthetic-filtered} &
  \multicolumn{1}{c|}{34.4 ± 0.23} &
  \multicolumn{1}{c|}{\textbf{51.98 ± 1.32}} &
  \multicolumn{1}{c|}{33.78 ± 0.22} &
  47.41 ± 2.17 \\ \hline
\end{tabular}}
\caption{Classification Results on IndicXNLI using 10-shot prompting.}
\label{tab:my-table}
\end{table}

\section{Training and Evaluation}
\label{sec:training hyperparameters appendix}
In this section, we provide an overview of the training and evaluation setup employed in our experiments. This includes details about the datasets used, training hyperparameters, evaluation metrics, and other relevant configurations.

\subsection{Evaluation Datasets}
\label{subsec: evaluation datasets appendix}
For evaluation, we utilize a diverse set of datasets covering four languages: English, Hindi, Gujarati, and Marathi. For Hindi and Gujarati, we rely on the IndicGLUE benchmark\footnote{\url{https://huggingface.co/datasets/ai4bharat/indic_glue}} \cite{kakwani-etal-2020-indicnlpsuite}, which provides a range of tasks for natural language understanding (NLU), including natural language inference (IndicXNLI/iXNLI), article genre classification (bbc-a, iNLTK), discourse mode classification (MIDAS), and sentiment analysis (iitp-mr, iitp-pr). For natural language generation (NLG), we employ the IndicNLG benchmark\footnote{\url{https://huggingface.co/collections/ai4bharat/indicnlg-66c5a1397bab135be074cfe1}} \cite{kumar-etal-2022-indicnlg}, which includes tasks like headline generation, sentence summarization, question generation, and Wikipedia biography generation.

The IndicGLUE dataset is semi-automatically curated using website metadata and Wikipedia articles, while IndicNLG is derived from Wikipedia articles and news websites for summarization, along with parallel corpora and pivot-based translation for paraphrasing tasks. Additionally, we incorporate the test sets from IN22 \cite{gala2023indictrans} and Flores-200 \cite{nllbteam2022language} to evaluate performance on machine translation tasks.

We use the well-known GLUE benchmark, which includes nine NLU tasks in English such as natural language inference (NLI), semantic similarity, text classification, and linguistic acceptability. For English summarization tasks, we rely on XLSum \cite{hasan-etal-2021-xl}, DialogSum \cite{chen-etal-2021-dialogsum}, and CNN/DailyMail \cite{see-etal-2017-get}.

\subsection{Training}
For the pretraining of the base models, we keep a hard limit for the \textit{base-1k} model as 2.38B tokens and for the \textit{mini-1k} model as 1B tokens. But for the TinyLM we relax this token limit until we see overfitting. For our experiments, we use the NVIDIA A100-SXM4-80GB GPUs.
\begin{table}[htbp!]
\centering
\resizebox{0.6\columnwidth}{!}{%
\begin{tabular}{c|c}
\hline
{\color[HTML]{333333} \textbf{Hyperparameter}} & \textbf{Value}         \\ \hline
{\color[HTML]{333333} vocab\_size}             & 56000                  \\
{\color[HTML]{333333} val\_every}              & 0.05                   \\
{\color[HTML]{333333} bs}                      & 48                     \\
{\color[HTML]{333333} n\_embed}                & 768                    \\
{\color[HTML]{333333} num\_blocks}             & 4                      \\
{\color[HTML]{333333} num\_heads}              & 16                     \\
{\color[HTML]{333333} head\_size}              & n\_embed // num\_heads \\
context\_len                                   & 1024                   \\
block\_size                                    & context\_len           \\
attn\_drop\_value                              & 0.1                    \\
dropout                                        & 0.1                    \\
ffn\_drop\_value                               & 0.1                    \\
use\_flashattn                                 & TRUE                   \\
ffn\_scaling                                   & 4                      \\
positional\_embedding                          & rope'                  \\
rotatory\_embedding\_dim                       & head\_size // 2        \\
lr                                             & 6.00E-04               \\
wd                                             & 1.00E-01               \\
beta\_1                                        & 0.9                    \\
beta\_2                                        & 0.95                   \\
eps                                            & 1.00E-05               \\
epochs                                         & 2                      \\
precision                                      & bf16                   \\
accumulate\_grad\_batches                      & 8                      \\
gradient\_clip\_val                            & 1                      \\
strategy                                       & ddp'                   \\
accelerator                                    & gpu'                   \\
warmup\_steps                                  & 5000                   \\
num\_workers                                   & 16                     \\
SHUFFLE\_SEED                                  & 42                     \\
PIN\_MEMORY                                    & TRUE                   \\
NUM\_\_NODES                                   & 1                      \\
NUM\_DEVICES                                   & 2                     
\end{tabular}
}
\caption{Hyperparameters used for training the \textit{mini-1k} model}
\label{tab:mini-1k hyperparameters}
\end{table}

\begin{table}[htbp!]
\centering
\resizebox{0.6\columnwidth}{!}{%
\begin{tabular}{c|c}
\hline
{\color[HTML]{333333} \textbf{Hyperparameter}} & \textbf{Value}         \\ \hline
{\color[HTML]{333333} vocab\_size}             & 56000                  \\
{\color[HTML]{333333} val\_every}              & 0.05                   \\
{\color[HTML]{333333} bs}                      & 48                     \\
{\color[HTML]{333333} n\_embed}                & 768                    \\
{\color[HTML]{333333} num\_blocks}             & 12                     \\
{\color[HTML]{333333} num\_heads}              & 12                     \\
{\color[HTML]{333333} head\_size}              & n\_embed // num\_heads \\
context\_len                                   & 1024                   \\
block\_size                                    & context\_len           \\
attn\_drop\_value                              & 0.1                    \\
dropout                                        & 0.1                    \\
ffn\_drop\_value                               & 0.1                    \\
use\_flashattn                                 & TRUE                   \\
ffn\_scaling                                   & 4                      \\
positional\_embedding                          & rope'                  \\
rotatory\_embedding\_dim                       & head\_size // 2        \\
lr                                             & 6.00E-04               \\
wd                                             & 1.00E-01               \\
beta\_1                                        & 0.9                    \\
beta\_2                                        & 0.95                   \\
eps                                            & 1.00E-05               \\
epochs                                         & 2                      \\
precision                                      & bf16                   \\
accumulate\_grad\_batches                      & 8                      \\
gradient\_clip\_val                            & 1                      \\
strategy                                       & ddp'                   \\
accelerator                                    & gpu'                   \\
warmup\_steps                                  & 5000                   \\
num\_workers                                   & 16                     \\
SHUFFLE\_SEED                                  & 42                     \\
PIN\_MEMORY                                    & TRUE                   \\
NUM\_\_NODES                                   & 1                      \\
NUM\_DEVICES                                   & 2                     
\end{tabular}
}
\caption{Hyperparameters used for training the \textit{base-1k} model}
\label{tab:base-1k hyperparameters}
\end{table}

\subsection{Extended pretraining}
For the \textit{mini-1k} models, we randomly sample 100M tokens from the clean subset of IndicMonoDoc for the target language, and for the \textit{base-1k} model, we sample 200M for extended pretraining. We use the same hyperparameters as training and perform extended pretraining for 2 epochs over this newly sampled clean data. For scaling experiments, we utilize TorchTune\footnote{\url{https://github.com/pytorch/torchtune}} for fine-tuning Llama-3-8B and Gemma-2B models. For a fair comparison, we limit each data split to 344M tokens for Gujarati and 465M tokens for Marathi and follow a similar procedure as described in Section \ref{sec:datafiltering} to generate and filter data for Marathi. We perform extended training for a single epoch using LoRA \citep{DBLP:journals/corr/abs-2106-09685} finetuning on $W_q$, $W_v$ projection matrices using $\alpha$=$16$ and $r$=$8$. We keep the learning rate at $3e^{-5}$ with a weight decay of $0.01$ and an effective batch size of 58k. We use the AdamW optimizer \cite{DBLP:conf/iclr/LoshchilovH19} with 1000 warmup steps and a cosine learning rate scheduler.

\subsection{Fine-tuning}
\label{sec: finetuning subsec appendix}
For GLUE tasks we use the dev split on the clean part and do hyperparameter tuning to achieve the best scores, and then we use the same hyperparameters for all synthetic experiments. For IndicGLUE we follow a similar setting for the val split to find good hyperparameters and report results on the test split like \citet{kakwani2020indicnlpsuite}. For all classification and regression tasks, we use a single linear layer and use an appropriate activation function for classification and regression respectively. We use an Adam optimizer \cite{DBLP:journals/corr/KingmaB14} with a learning rate of $1e^{-5}$ and a batch size of 48. For NLG tasks we do extended pretraining using a separator token in between the input and output sequence with an effective batch size of 768 examples and only calculate loss for the output sequence. We use an AdamW optimizer \cite{DBLP:conf/iclr/LoshchilovH19} with learning rate = $6e^{-4}$, weight decay = $1e^{-1}$, $\beta_1$ = 0.9, $\beta_2$ = 0.95 and $\epsilon$ = $1e^{-5}$. For translation, we randomly sample 1M parallel sentence for each language pair from the samanantar corpus \cite{ramesh-etal-2022-samanantar} and evaluate on FloRes \cite{nllbteam2022language}, IN22-Conv and IN22-Gen \cite{gala2023indictrans}. We list the batch size and number of epochs of each task in Table \ref{tab:finetuning hyperparameters appendix}.

\subsection{Prompting}
\label{app: prompts}
We use random sampling from validation sets whenever available and utilize other examples from the test set otherwise. We take 5 random samples for each evaluation of NLG tasks and 10 random samples for each evaluation of NLU tasks. We list down the prompt used below for Marathi evaluations, we use similar prompts for Gujarati as well.

\begin{figure}[h!]
    \centering
    \includegraphics[width=0.45\textwidth]{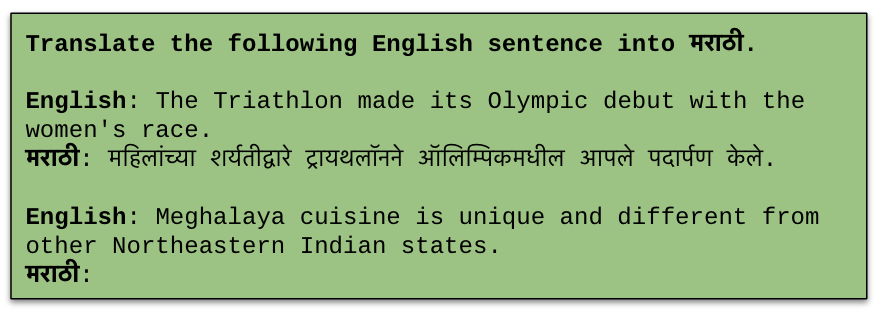}
    \caption{Prompt used for English$\rightarrow$Marathi translation.}
    \label{fig: prompt}
\end{figure}

\begin{table}[htbp!]
\centering
\resizebox{0.9\columnwidth}{!}{%
\begin{tabular}{c|ccc}
\hline
{\color[HTML]{333333} \textbf{Task}}          & \textbf{Batch size} & \textbf{Epochs} & \textbf{Metric} \\ \hline
{\color[HTML]{333333} IndicXNLI} & 48  & 5  & Accuracy      \\
{\color[HTML]{333333} BBC-Articles}           & 24                  & 20              & Accuracy        \\
{\color[HTML]{333333} IITP-MR}   & 24  & 20 & Accuracy      \\
{\color[HTML]{333333} IITP-PR}   & 48  & 20 & Accuracy      \\
{\color[HTML]{333333} MIDAS}     & 48  & 20 & Accuracy      \\
{\color[HTML]{333333} Headline Generation}    & 768                 & 2               & Rouge-L F1      \\
{\color[HTML]{333333} Sentence Summarization} & 768                 & 2               & Rouge-L F1      \\
Question Generation              & 768 & 2  & Rouge-L F1    \\
WikiBio Generation               & 768 & 4  & Rouge-L F1    \\
iNLTK                            & 48  & 20 & Accuracy      \\
sst2                             & 48  & 10 & Accuracy      \\
CoLA                             & 48  & 30 & MCC           \\
mrpc                             & 48  & 30 & F1            \\
qnli                             & 48  & 10 & Accuracy      \\
qqp                              & 48  & 5  & F1            \\
rte                              & 48  & 30 & Accuracy      \\
mnli-matched                     & 48  & 5  & Accuracy      \\
mnli-mismatched                  & 48  & 5  & Accuracy      \\
stsb                             & 48  & 20 & Pearson       \\
XLSum Headline Gen.              & 768 & 4  & Rouge-L F1    \\
XLSum Question Gen.              & 768 & 4  & Rouge-L F1    \\
CNN Dailymail                    & 768 & 4  & Rouge-L F1    \\
DialogSum                        & 768 & 4  & Rouge-L F1    \\
Samanantar                       & 768 & 2  & chrF++ / BLEU
\end{tabular}
}
\caption{Hyperparameters used for finetuning tasks}
\label{tab:finetuning hyperparameters appendix}
\end{table}

\subsection{Evaluation}
We use torch metrics\footnote{\url{https://lightning.ai/docs/torchmetrics/stable/pages/lightning.html}} to calculate accuracy, f1-score, Pearson correlation, Matthew's correlation coefficient. We report chrF++ scores\footnote{chrF++ signature\\nrefs:1|case:mixed|eff:yes|nc:6|nw:2|space:no|version:2.4.0} and BLEU scores\footnote{sacreBLEU signature:\\nrefs:1|case:mixed|eff:no|tok:13a|smooth:exp|version:2.4.0} \cite{papineni-etal-2002-bleu} using the sacreBLEU\footnote{\url{https://github.com/mjpost/sacrebleu}} implementation and Rouge-L f1 scores using the sacreRouge \cite{deutsch-roth-2020-sacrerouge} implementation by the xl-sum repository\footnote{\url{https://github.com/csebuetnlp/xl-sum}}. 

We report English scores for NLU on the validation split of the GLUE benchmark and test splits for XL-Sum, CNN Dailymail, and Dialogsum NLG benchmarks. For Hindi and Gujarati, we use the test split of IndicGLUE and IndicXNLI.

For classification and regression tasks, we use the models finetuned according to hyperparameters mentioned in Appendix \ref{sec: finetuning subsec appendix} to keep fair comparison for all models and mention results on the final epoch.
For generations on IndicNLG and English NLG tasks, we use beam search with a beam width of 5, length penalty of 1.0, n\_gram repetition penalty of 4 n\_grams with sampling set to false and early stopping set to true. We also set a maximum generation length to 64 tokens. For the translation task, we follow a beam search with a beam width of 5, maximum new tokens to 256 and early stopping to true. 

\section{Utilising Translationese}
\label{sec:document filtering appendix}

In this section we provide details on the generation and filtering of translationese data for our experiments.

\subsection{Creating synthetic data}
``Translationese'' is a term used to describe peculiarities in the text translated into a specific language, differentiating it from content originally written in that language \cite{gellerstam1986translationese}. Translated texts into the target language (via humans or machine-generated) often show distinctive features that differentiate them from their original counterparts in the target language. These disparities arise from either the influence of the translation process itself on the final product or the inherent ``fingerprints'' of the source language subtly present in the target language rendition \cite{rabinovich2015unsupervised}. This is a common phenomenon in translation models where the target language translations often show characteristics of the source language and add bias to the evaluation of downstream tasks \cite{toral-etal-2018-attaining, DBLP:journals/corr/abs-1906-08069, DBLP:journals/corr/abs-1906-09833}. So far a lot of work on synthetic translated data has been done for using back translations \cite{sennrich-etal-2016-improving, edunov-etal-2018-understanding} for improving Machine translation performance \cite{marie-etal-2020-tagged, DBLP:journals/corr/abs-1911-03362, ni-etal-2022-original} or for classification tasks like native language identification \cite{goldin-etal-2018-native}, etc. Tranlationese data has been used for many tasks but we explore the efficacy of using translationese data for pretraining of language models. We collect monolingual corpora in the source language as mentioned in Section \ref{subsec: Indicmonodoc} and utilize a powerful off-the-shelf translation model IndicTrans2 \cite{gala2023indictrans} to generate translationese data. Since IndicTrans2 can only handle a max sentence length of 256 BPE tokens, we split the documents using Moses Sentence Splitter\footnote{\url{https://pypi.org/project/mosestokenizer/}} to perform translations into the target language at the sentence level and then merge again to form documents. We also repair translations that exceed in length 256 BPE tokens using the TinyLM trained on clean corpora as mentioned in Section \ref{sec: experiments} to complete the sentence translation, we encounter only 0.002\% of such cases. We use this corpus for the \textit{synthetic} and \textit{clean+synthetic} part of our experiments.

\begin{figure}[h!]
    \centering
    \includegraphics[width=0.49\textwidth]{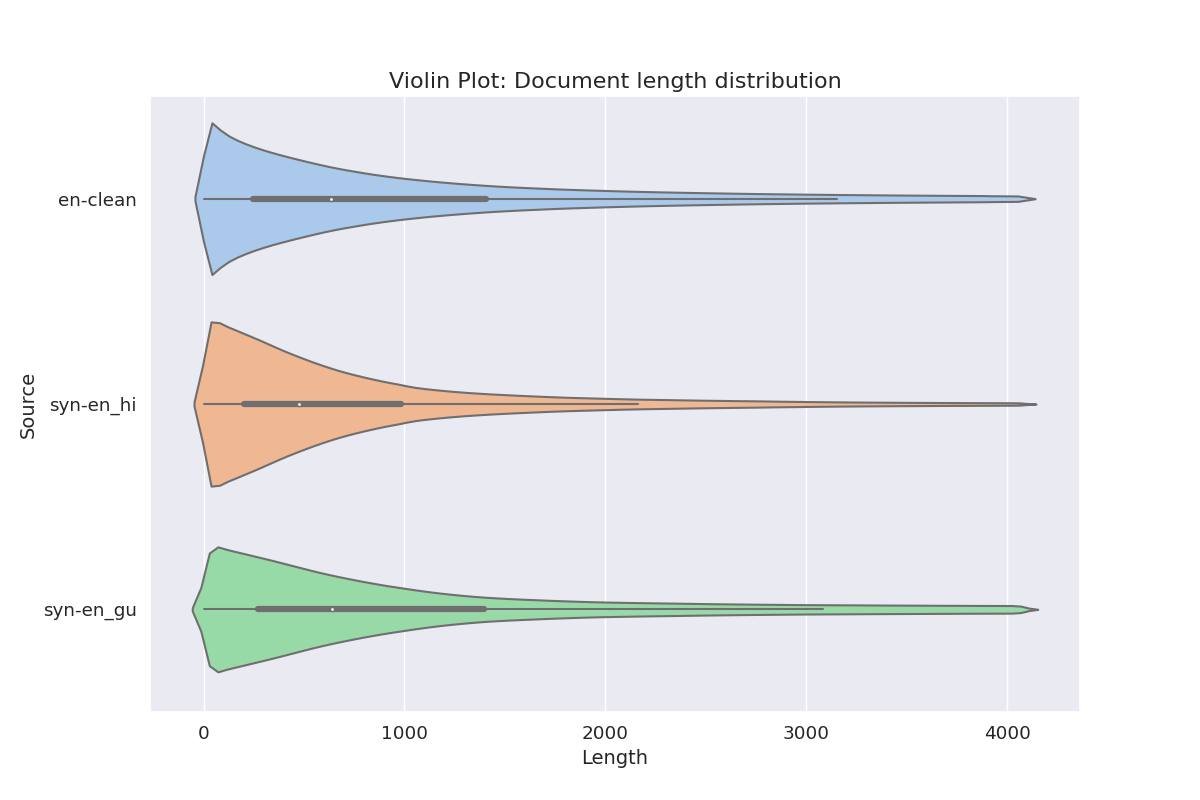}
    \caption{Violin plot displaying the distribution of lengths of clean and filtered English documents on different data splits: \textit{en-clean} (English web documents), \textit{syn-en\_hi} (synthetic English documents translated from Hindi), and \textit{syn-en\_gu} (synthetic English documents translated from Gujarati).}
    \label{fig: violin plot}
\end{figure}

\subsection{Perplexity filtering}
Following Figure \ref{fig:filtering-translationese}, we use these TinyLMs to filter the generated synthetic translationese corpora from IndicTrans2. We do this by using perplexity as a measure of document quality score. For language models, perplexity quantifies how well a model predicts a sequence of tokens. A lower perplexity indicates better predictive performance. While calculating perplexity over a sequence of tokens $W \in$ \( w_1, w_2, \ldots, w_N \) we skip the first $s$ tokens where $s=10$, $e=1024$ and calculate loss until only the first $e$ tokens of the document. We find setting $e$ to larger values can lead to higher variance in the document scores due to the size of the TinyLM. After initial analysis, we choose $s$ and $e$ such that we remove the high uncertainty of the language at the start of an unseen document and avoid penalizing longer documents due to the fragility of the extrapolation ability of TinyLM\footnote{During experiments we saw that these TinyLMs can only go up to a certain context length before deteriorating in quality.}. Note that it is important to choose $e$ such that the language model gives a uniform estimate of perplexity over an already seen sequence of tokens $\in$ {\( w_s, w_{s+1}, \ldots, w_e \)}. For our experiments, we use the TinyLMs to score all synthetically generated translationese data and calculate a document score using the above method. Following \citet{laurenccon2022bigscience}, we do subsampling by thresholding document perplexity scores except \citet{laurenccon2022bigscience} did it using Ken-LM \cite{heafield-2011-kenlm} and we do it using our TinyLM. We keep the threshold value such that we include enough documents to reach the computed optimal token count for pretraining experiments.

\begin{table}[]
\centering
\resizebox{0.8\columnwidth}{!}{%
\begin{tabular}{|cc|cc|}
\hline
\multicolumn{2}{|c|}{\textbf{High Resource}}            & \multicolumn{2}{c|}{\textbf{Low Resource}}                 \\ \hline
\multicolumn{1}{|c|}{\textbf{Lang}} & \textbf{\#Tokens} & \multicolumn{1}{c|}{\textbf{Lang}} & \textbf{\#Tokens}     \\ \hline
\multicolumn{1}{|c|}{bn} & 5,258.47  & \multicolumn{1}{c|}{as}  & 57.64 \\ \hline
\multicolumn{1}{|c|}{en} & 11,986.53 & \multicolumn{1}{c|}{brx} & 2.25  \\ \hline
\multicolumn{1}{|c|}{gu} & 887.18    & \multicolumn{1}{c|}{doi} & 0.37  \\ \hline
\multicolumn{1}{|c|}{hi} & 11,268.33 & \multicolumn{1}{c|}{gom} & 2.91  \\ \hline
\multicolumn{1}{|c|}{kn} & 567.16    & \multicolumn{1}{c|}{kas} & 1.27  \\ \hline
\multicolumn{1}{|c|}{ml} & 845.32    & \multicolumn{1}{c|}{mai} & 1.51  \\ \hline
\multicolumn{1}{|c|}{mr} & 1,066.76  & \multicolumn{1}{c|}{mni} & 0.99  \\ \hline
\multicolumn{1}{|c|}{ne} & 1,542.39  & \multicolumn{1}{c|}{or}  & 81.96 \\ \hline
\multicolumn{1}{|c|}{pa} & 449.61    & \multicolumn{1}{c|}{sa}  & 80.09 \\ \hline
\multicolumn{1}{|c|}{ta} & 2,171.92  & \multicolumn{1}{c|}{sat} & 3.05  \\ \hline
\multicolumn{1}{|c|}{te} & 767.18    & \multicolumn{1}{c|}{sd}  & 83.81 \\ \hline
\multicolumn{1}{|c|}{ur}            & 2,391.79          & \multicolumn{1}{l|}{}              & \multicolumn{1}{l|}{} \\ \hline
\end{tabular}
}
\centering
\caption{Languagewise corpora size in Million tokens}
\label{table 1: corpora comparison}
\end{table}

\section{Qualitative Analysis}
\label{sec:qualitative appendix}

Since translation errors occur frequently, leading to biased, ungrammatical, or erroneous translations that can have drastic consequences on training, it is important to mitigate or remove such errors in translationese corpora. The most common machine translation errors include mistranslations due to ambiguous words, incorrect handling of expressions, syntax and grammar errors, and issues with preserving context across longer sentences. Many approaches have been proposed to address these errors, but most are computationally expensive, especially when translationese data is used for pretraining. Instead, we ask whether a language model can identify such issues. To investigate this, we examine which types of English sentences were filtered.

\begin{figure}[h!]
    \centering
    \fbox{\includegraphics[width=0.45\textwidth]{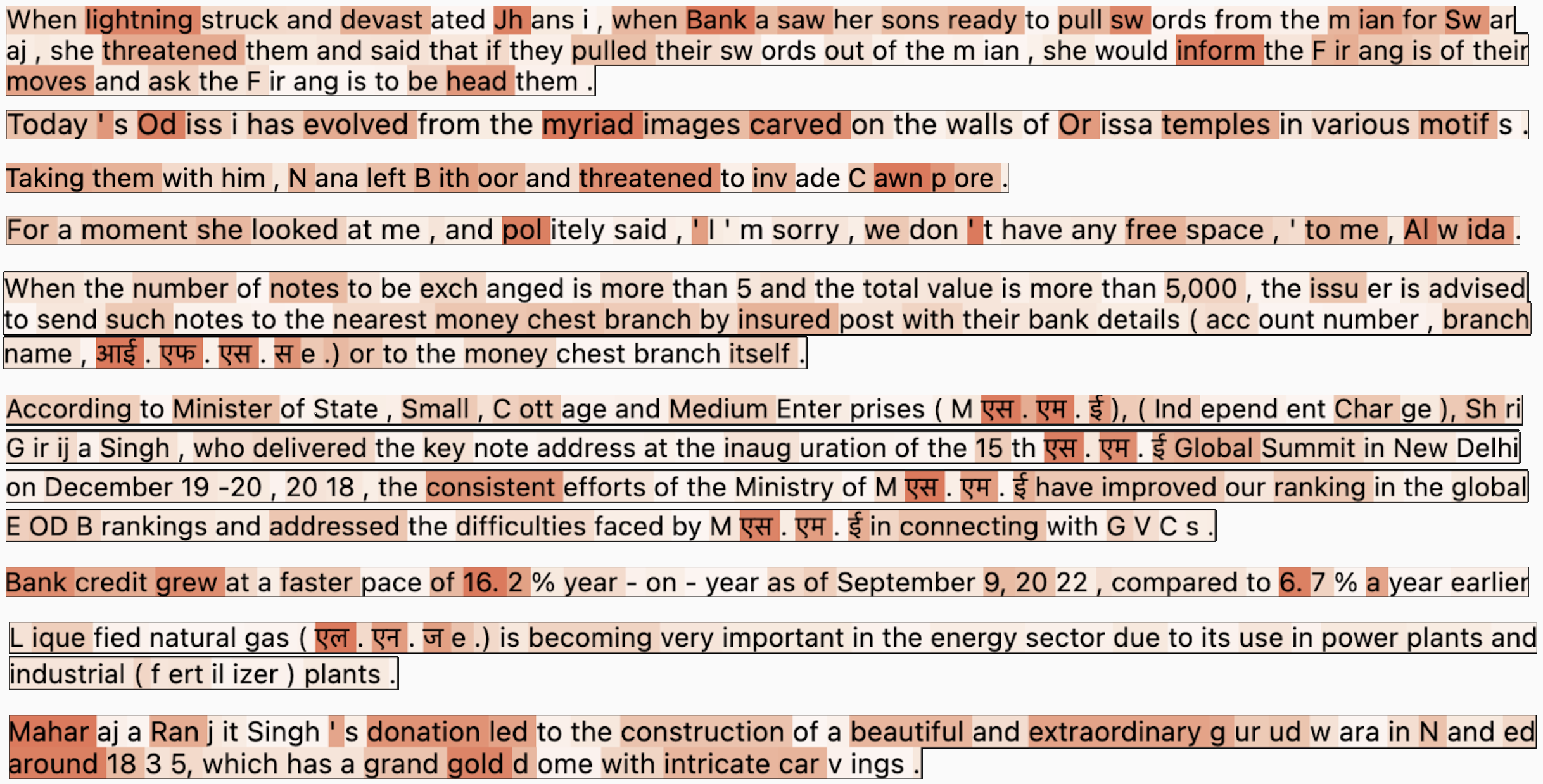}}
    \caption{Heatmap of perplexity over filtered sentences.}
    \label{fig: qualitative examples filtered}
\end{figure}

In many cases, we found that the filtered documents included errors like code-mixing and repetitions, often generated by the diverging output of the translation model. This is expected since such phenomena are rarely seen in natural written language, and the model assigning high entropy suggests an unlikely sequence outcome. Although the model regarded some erroneous instances as false positives, many such cases were successfully avoided. We also noticed that, due to the small size of the model, many perfectly good documents were filtered out because of the model's inability to evaluate them. This issue was observed less frequently in larger models used for evaluation. As seen in Figure \ref{fig: qualitative examples filtered}, the filtering model fails to understand complex words and named entities, which it regarded as unlikely due to its preference for simpler terms and more likely entities. Other common elements that were filtered included numbers, dates, and abbreviations. While this can lead to the loss of valuable information, as many good documents are discarded, we empirically observe the benefits of such filtering. These errors could likely be reduced by using a larger filtering model that can better approximate the source language but we leave this analysis for future work.

\section{IndicMonoDoc}
\label{sec:indicmonodoc appendix}

In this section, we describe the process of creating the IndicMonoDoc corpus which is the largest document-level corpora for Indic languages consisting of 39.5 billion tokens spanning 23 languages. IndicMonoDoc comprises 27.5B Indic tokens and 12B tokens of English tokens. Table \ref{table 1: corpora comparison} shows language language-wise deduplicated size of the IndicMonoDoc corpus and Figure \ref{fig:IndicMonoDoc} shows a comparative 100\% stacked bar plot with IndicCorpv2 which is a sentence level corpora.

\subsection{Crawling}
To extract URLs from the web we sample word level \textit{n-grams}; \textit{n={2,...,6}} from a sample monolingual corpora to create a list of keyword searches. We then randomly merge \textit{k}; \textit{k={1,..,4}} keywords to form a query. Using these queries we perform automatic web searches to collect a large repository of URLs. We merge this list with a manual list of sources to perform URL-level deduplication. We crawl these webpages leaving out some of them\footnote{We leave webpages consisting of a robots.txt file and URLs containing offensive text or social media links}. We leave out webpages that consist of a considerable amount of English content using a simple script recognition regex. We perform this scrapping majorly for the bottom 14 low-resource languages. We also add script-level recognition using Unicode characters\footnote{\url{https://unicode.org/charts/}} for each language before crawling a webpage to avoid scrapping non-Indic text.

\subsection{Post processing}
A lot of crawled content consists of unwanted text like HTML tags, emoticons, and text in another language. We use manual filtering pipelines inspired by OSCAR \cite{OrtizSuarezSagotRomary2019}, \cite{2022arXiv220106642A} to remove such content. We additionally use a language detection-based (LID) filtering using cld3\footnote{\url{https://github.com/google/cld3}} and IndicLID-FTN model \cite{madhani2023bhashaabhijnaanam} to discard languages not of interest. Following \citet{doddapaneni2023towards} we perform document filtering to remove offensive text from the corpora using a list of offensive words and phrases extended from work by \citet{nllbteam2022language} which consists of offensive words in 209 languages. We also use a Romanized version of this list using the transliteration tool by \citet{madhani-etal-2023-aksharantar} to perform toxic document filtering in 17 languages. Following \citet{kakwani2020indicnlpsuite} \& \citet{doddapaneni2023towards} we merge all the filtered corpus with Wikipedia, OSCAR \cite{OrtizSuarezSagotRomary2019} and some dumps of mC4 \cite{xue-etal-2021-mt5}. Finally, we perform deduplication at paragraph level using Murmurhash algorithm\footnote{\url{https://pypi.org/project/mmh3/}} with a 128-bit unsigned hash for each monolingual split of the corpora. After all post-processing steps, the language wise size of the corpora is mentioned in Table \ref{table 1: corpora comparison}. A major chunk of the corpus is comprised of English, Hindi, and Bengali which make up 72.15\% of the corpora.

\end{document}